\documentclass[10pt,twocolumn,letterpaper]{article}
\usepackage[pagenumbers]{cvpr}

\definecolor{cvprblue}{rgb}{0.21,0.49,0.74}
\usepackage[pagebackref,breaklinks,colorlinks,allcolors=cvprblue]{hyperref}
\usepackage{booktabs}
\usepackage{multirow}
\usepackage{makecell}
\usepackage{graphicx}
\usepackage{colortbl}
\usepackage{xcolor}

\title{ACoT-VLA: Action Chain-of-Thought for Vision-Language-Action Models}

\author{
Linqing Zhong$^{1, 2}$~\quad Yi Liu$^{2}$~\quad Yifei Wei$^{1, 2}$~\quad Ziyu Xiong$^{2}$ \\
Maoqing Yao$^{2}$\thanks{Corresponding authors.}~\quad Si Liu$^{1}$\footnotemark[1]~\quad Guanghui Ren$^{2}$\footnotemark[1] \\[0.5em]
$^1$Beihang University~\quad $^2$AgiBot
}

\begin{document}
\maketitle
\begin{abstract}
Vision-Language-Action models have emerged as essential generalist robot policies for diverse manipulation tasks, conventionally relying on directly translating multimodal inputs into actions via Vision-Language Model embeddings.
Recent advancements have introduced explicit intermediary reasoning—such as sub-task prediction (language) or goal image synthesis (vision)—to guide action generation.
However, these intermediate reasoning are often indirect and inherently limited in their capacity to convey the full, granular information required for precise action execution.
Instead, we posit that the most effective form of reasoning is one that deliberates directly in the action space.
We introduce \textbf{Action Chain-of-Thought (ACoT)}, a paradigm where the reasoning process itself is formulated as a structured sequence of coarse action intents that guide the final policy. 
In this paper, we propose ACoT-VLA, a novel architecture that materializes the ACoT paradigm.
Specifically, we introduce two complementary components: an Explicit Action Reasoner (EAR) and Implicit Action Reasoner (IAR).
The former proposes coarse reference trajectories as explicit action-level reasoning steps, while the latter extracts latent action priors from internal representations of multimodal input, co-forming an ACoT that conditions the downstream action head to enable grounded policy learning.
Extensive experiments in real-world and simulation environments demonstrate the superiority of our proposed method. Code is available at: \url{https://github.com/AgibotTech/ACoT-VLA}.
\end{abstract}    
\section{Introduction}
\label{sec:intro}
\begin{figure}[t]
  \includegraphics[width=\columnwidth]{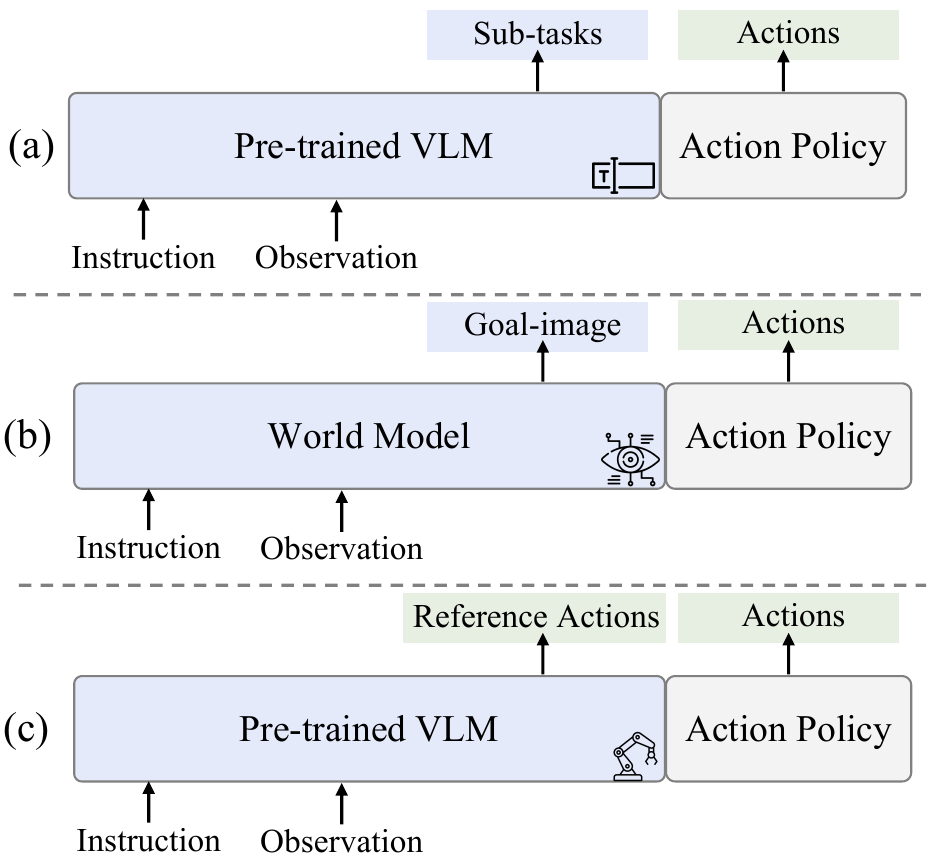}
  \vspace{-5.5mm}
  \caption{Chain-of-Thought in different space. (a) Language CoT paradigm predicts sub-tasks as intermediate reasoning. (b) Visual CoT paradigm synthesizes a goal image to provide guidance for action policy. (c) Our proposed Action CoT directly operates in action space and provides homogeneous action guidance.}
  \vspace{-4mm}
  \label{fig:fig1}
\end{figure}

To overcome the generalization limits of task-specific robot policies~\cite{chi2025diffusion, zhao2023learning, team2024octo}, recent work has converged on Vision-Language-Action (VLA) models~\cite{kim2024openvla, black2410pi0, qu2025spatialvla, liu2024rdt}, which always leverage a pre-trained Vision-Language Model (VLM)~\cite{achiam2023gpt, team2023gemini, bai2025qwen2} to encode visual and linguistic inputs into a latent representation that conditions an action decoder.
Recent advancements seek to improve the mapping from the input space to the action space by introducing the intermediate reasoning step by language generation, leading to more generalized and precise action outputs~\cite{intelligence2025pi_, zawalski2024robotic}, as visualized in Fig.~\ref{fig:fig1}(a).
A parallel thrust leverages world models~\cite{ha2018world, yan2025renderworld, zheng2024occworld} to simulate environmental dynamics, directly enhancing the efficacy and goal-oriented nature of the generated action sequences~\cite{zhen20243d, zhang2025dreamvla}, as shown in Fig.~\ref{fig:fig1}(b).

Despite the promising trajectory set by these paradigms, a critical challenge persists: existing generalist policies think predominantly in the vision-language (input) space, often failing to adequately address the inherent disparity between these rich, semantic representations and the requirements of precise, low-level action execution (output). 
Specifically, the knowledge encoded within the VLM backbone of VLA models is derived from pre-training on web-scale datasets focused on semantic alignment and question-answering, yielding representations optimized for linguistic understanding rather than physical dynamics. 
Similarly, while world models forecast future visual states conditioned on inputs, their guidance remains tethered to naturally visual representations. 
Crucially, both semantic and visual forms of reasoning only offer suboptimal, indirect guidance for generating the necessary action sequence.
Consequently, these prevailing approaches rely on an inherently constrained information conduit, struggling to convey the full, granular knowledge of the action space essential for truly grounded and accurate robotic policy learning.

The inherent semantic-kinematic gap in existing policies, \ie, a fundamental disconnect between high-level, abstract inputs and low-level, executable motor commands, necessitates a paradigm shift in how guidance is provided.
We contend that to bridge this chasm, policies require guidance that is kinematically coherent, rather than purely semantic or visual. 
This core principle underpins our novel framework: \textbf{Action Chain-of-Thought (ACoT)} (Fig. ~\ref{fig:fig1}(c)).
We redefine the ``thought'' process not as a sequence of linguistic tokens, but as a structured chain of explicit, kinematically-grounded action intents. 
This approach furnishes the policy with direct motion cues, supplanting indirect representations. 
In a manner analogous to learning from physical demonstration, this direct conditioning on action-space information enables a substantially more efficient and veridically grounded policy learning process.

This foundational shift, however, introduces a critical and distinct research challenge: \textit{``How can we robustly and efficiently synthesize the complex, high-dimensional motion cues required for ACoT reasoning from the raw, heterogeneous multimodal inputs?''}

Action-related information manifests in two complementary forms, \ie, explicit or implicit. 
The explicit form corresponds to observable motion trajectories, such as those in human demonstrations, which directly encode executable patterns of behavior. 
In contrast, the implicit form resides in latent cues, \eg, linguistic expressions like ``reach out'' or ``grasp'', as well as interaction intents embedded in visual contexts. 
Although these cues are not presented as explicit robotic trajectories, they implicitly define distributions over feasible actions within the action space.
Building upon this insight, we introduce two synergistic mechanisms to generate both explicit and implicit guidance in the action space. 
We first propose the Explicit Action Reasoner (EAR), which is realized as a light-weight transformer. Particularly, EAR synthesizes coarse-grained motion trajectories conditioned on multimodal observations, offering direct and executable guidance within the action space. 
Secondly, we devise the Implicit Action Reasoner (IAR), which infers latent action priors through applying cross-attention modeling between downsampled multimodal representations and learnable queries, thereby providing implicit behavioral priors.
Note that these two mechanisms are inherently complementary to each other. 
Subsequently, through jointly leveraging both EAR and IAR, we develop ACoT-VLA, an integrated Action Chain-of-Thought framework that enables grounded generalist robot policy learning. 
Extensive experiments across both real-world settings and three simulation benchmarks consistently demonstrate the effectiveness and versatility of our ACoT-VLA.

To summarize, our main contributions are as follows:
\begin{itemize}
    \item Conceptually, we introduce Action Chain of Thought (ACoT), a new paradigm for generalist robot policies. To the best of our knowledge, this is the first work to formulate the deliberative process as a structured chain of explicit action-space intents, rather than abstract linguistic or visual sub-goals.
    \item We delve into essential action space guidance and propose the Explicit and Implicit Action Reasoners, which provide both explicit trajectory guidance and implicit behavioral inspiration for action prediction.
    \item Building upon these two modules, we further propose ACoT-VLA, a unified framework for grounded generalist robot policy learning.
    \item Empirically, we validate our approach through extensive simulation and real-world experiments, achieving state-of-the-art performance on multiple benchmarks.
\end{itemize}

\section{Related Works}
\label{sec:works}

\begin{figure*}[t]
  \includegraphics[width=\linewidth]{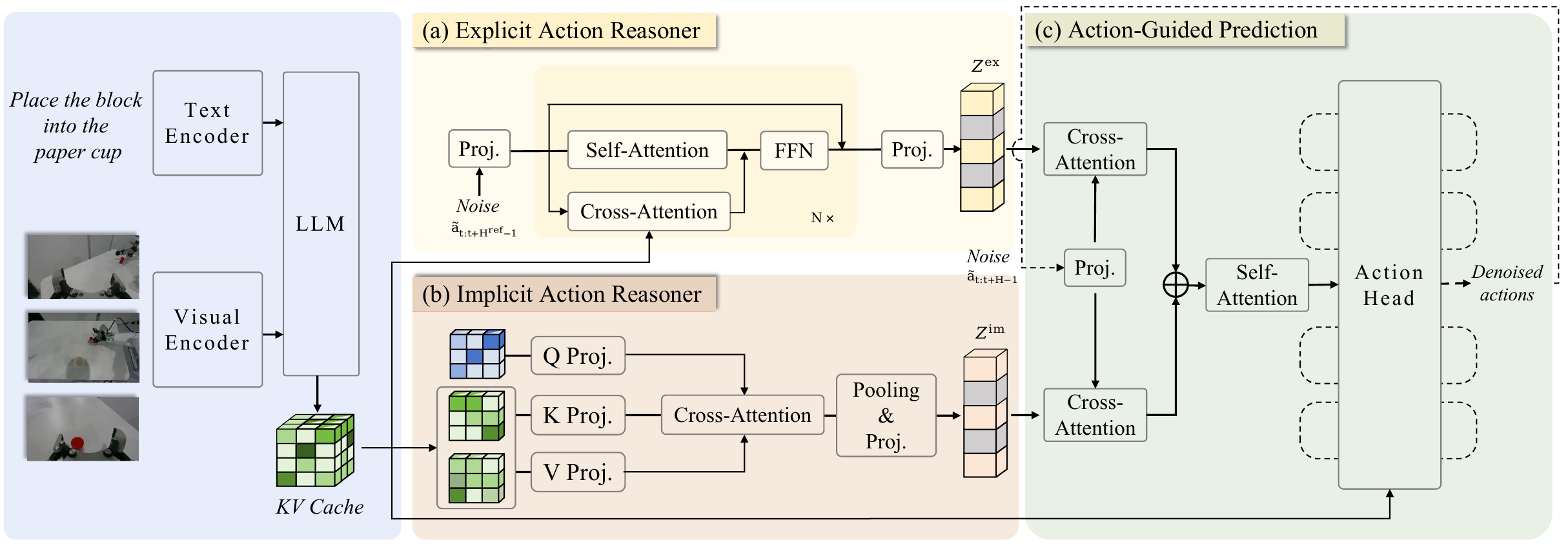}
  \vspace{-5.5mm}
  \caption{{Architectural Overview of ACoT-VLA.
    The framework consists of three main components operating on features from a shared VLM backbone.
    (a) The Explicit Action Reasoner (EAR) is a Transformer-based module that synthesizes a coarse reference trajectory, providing explicit action-space guidance.
    (b) The Implicit Action Reasoner (IAR) employs a cross-attention mechanism with learnable queries to extract latent action priors from the VLM's internal representations.
    (c) The Action-Guided Prediction (AGP) head synergistically integrates both explicit and implicit guidances via cross-attention to condition the final denoising process, producing the executable action sequence.}}
  \vspace{-4mm}
  \label{fig:fig2}
\end{figure*}

\noindent\textbf{Vision-Language-Action Models.}
VLA models~\cite{duan2024manipulate, huang2024a3vlm, huang2024rekep, li2024manipllm} incorporate pre-trained VLM models to predict language-driven robotic action sequences. Early works~\cite{zitkovich2023rt, kim2024openvla} formulate robot control as an autoregressive sequence generation problem, discretizing continuous actions into bins. Inspired by generative modeling~\cite{peebles2023scalable, zhou2024transfusion, lipman2022flow}, increasing works~\cite{black2410pi0, liu2024rdt, huang2025motvla} adopt diffusion-based action policies to synthesize smooth and high-quality action trajectories. Given that robotic manipulation inherently occurs in three-dimensional space, a line of studies~\cite{yuan2025depthvla, lin2025evo, zhang2025spatial} have sought to enhance the spatial reasoning capability of VLA models by integrating 3D priors. For instance, SpatialVLA~\cite{qu2025spatialvla} integrates spatial embeddings to endow model with 3D awareness, while 4D-VLA~\cite{zhang20254d} incorporates both spatial and temporal information to enrich representations. 
Besides, due to the scarcity of large-scale real-world robot demonstrations, a series of efforts~\cite{o2024open, zhang2025vlabench, deng2025graspvla, bu2025agibot, jiang2025galaxea, chen2025robotwin} focus on data-centric solutions, constructing large-scale robotic datasets through simulation or real-world collection to scale up policy learning.
Moreover, recent large-scale co-training approaches, such as $\pi_{0.5}$~\cite{intelligence2025pi_}, GenieReasoner~\cite{liu2025unified} and Gemini Robotics~\cite{team2025gemini}, demonstrate the potential of unifying web-scale language understanding with action learning, enhancing the policy’s generalization ability while retaining the reasoning capability of pre-trained foundation models.

\vspace{0.5mm}
\noindent\textbf{World-Model-based Policies.}
Advances in world models have demonstrated remarkable capability in synthesizing high-fidelity images and temporally coherent videos. Building upon such progress, emerging researches~\cite{li2025vla, lv2025f1, zhang2025up, liao2025genie} exploit their predictive dynamics to implicitly guide action generation.
Specifically, CoT-VLA~\cite{zhao2025cot} introduces visual chain-of-thought reasoning by forecasting sub-goal images, explicitly integrating visual reasoning into action prediction. WorldVLA~\cite{cen2025worldvla} employs an autoregressive architecture that unifies perception and action generation within a single framework. DreamVLA~\cite{zhang2025dreamvla} extends beyond visual prediction and enriches world modeling with dynamic, depth, and semantic cues, improving the model’s physical consistency.
Collectively, existing world-model-based methods adopt knowledge-forecasting perspective, incorporating primarily visual guidance into trajectory generation.

In contrast to previous works focusing on visual or linguistic intermediaries for robotic policy learning, our key insight lies in investigating guidance directly within the action space, which intrinsically mitigates the heterogeneity between perception and action, enabling the model to effectively learn action-relevant priors.
\section{Methodology}
\label{sec:method}
In this section, we present a detailed investigation into how to generate effective action space guidance and integrate it into robotic policy learning. 
We first define the robotic manipulation problem and formulate our proposed approach in Sec.~\ref{sec:formulation}. 
The core of our method lies in two distinct action reasoners introduced in Sec.~\ref{sec:EAR} and Sec.~\ref{sec:IAR}, which provide explicit and implicit guidance within the action space. 
We conclude by illustrating the policy prediction strategy that effectively integrates this action guidance during policy learning (Sec.~\ref{sec:AGP}).

\subsection{Problem Formulation}
\label{sec:formulation}
Given a natural language instruction $l$ and current visual observation $o_t$, the generalist robot policy  $\pi_{\theta}$ aims to predict action sequences $a_{t:t+H-1}$ that accomplishes the specified task. The process can be formally expressed as:
\begin{equation}
    a_{t:t+H-1} = \pi_{\theta}(o_t, l),
\end{equation}
where $H$ represents the action horizon. Numerous works introduce additional guidance signals $g$, which encapsulates various forms of auxiliary information to enhance policy’s prediction ability.
Specifically, these guidance signals can be broadly categorized into two types: language-level guidance $g_{\text{lang}}$ and vision-level guidance $g_{\text{vis}}$. The former is primarily adopted by VLM-based methods, \eg, leveraging LLMs' reasoning capabilities to predict sub-tasks, while the latter is always employed by world-model-based approaches, such as simulating future observations.
Such relationship can be formulated as:
\begin{equation}
    \pi_{\theta}(a_{t:t+H-1}, g\mid o_t, l) = \pi_{\theta}(a_{t:t+H-1}\mid o_t, l, g)\pi_{\theta}(g\mid o_t, l),
\end{equation}
where $g \in \{ g_{\text{lang}},\, g_{\text{vis}} \}$. Conversely, we shift the focus toward the action domain itself and investigate cues operating directly in the action space, symbolized as $g_{\text{action}}$. The above guidances are extended as $g \in \{ g_{\text{lang}},\, g_{\text{vis}},\, g_{\text{action}} \}$.

Such action guidance can intuitively be disentangled into explicit and implicit forms. The explicit guidance $g_{\text{action}}^{\text{ex}}$ provides direct priors in the form of reference action sequences, whereas the implicit guidance $g_{\text{action}}^{\text{im}}$ arises from contextual signals, \eg, action distribution inherently implied in linguistics.

\subsection{Explicit Action Reasoner}
\label{sec:EAR}
To incorporate explicit action trajectories into the thinking process of  $\pi_{\theta}$ to generate high-quality action predictions, we propose the Explicit Action Reasoner (EAR).

We design a mechanism that enables the model to autonomously synthesize reference action sequences as internal guidance for policy learning.
Analogously, this formulation can be viewed as an action-space transfer of self-conditioning in generative models~\cite{chen2022analog, liu2022self}, where incorporating prior estimates into the generation process has been shown to markedly improve sample quality.
Building upon this principle, we instantiate EAR as a light-weight transformer, as shown in Fig.~\ref{fig:fig2} (a), generating kinematically plausible action reference as explicit action-space guidance $g_{\text{action}}^{\text{ex}}$ for downstream action policy.

Formally, given visual observation $o_t$ and language instruction $l$, a pre-trained VLM encodes them into a contextual key-value cache:
\begin{equation}
    (K^{\text{VLM}}_{1:N}, V^{\text{VLM}}_{1:N}) = \text{VLM}(o_t, l),
\end{equation}
where $N$ represents the number of layers of VLM.
Subsequently, the EAR, denoted as $\pi_\theta^{\text{ref}}$, takes a noisy action sequence $\tilde{a}_{t:t+H^{ref}-1}$ as input, where $H^{ref}$ indicates the horizon of reference actions. The sequence is first embedded into an initial hidden representation $h_0^{\text{ref}}$, which serves as the input to EAR's transformer layers. At each transformer layer $i$, we adopt self-attention, along with cross-attention with the contextual key-value cache from the corresponding VLM layer:
\begin{equation}
\tilde{h}_i^{\text{ref}} = \text{Self-Attn}(h_{i-1}^{\text{ref}}) + \text{CrossAttn}(h_{i-1}^{\text{ref}}, K^{\text{VLM}}_i, V^{\text{VLM}}_i),
\end{equation}
where self-attention module captures temporal dependencies within the action sequence and cross-attention mechanism injects multimodal contextual priors from the VLM.
Then, the intermediate representation $\tilde{h}_i^{\text{ref}}$ is processed by a feed-forward network (FFN) in a residual-parallel manner, updating the $i$-th EAR representation $h_i^{\text{ref}}$:
\begin{equation}
h_i^{\text{ref}} = h_{i-1}^{\text{ref}} + \text{FFN}(\tilde{h}_i^{\text{ref}}).
\end{equation}

Through training via flow matching, $\pi_\theta^{\text{ref}}$ learns a distribution over action trajectories, producing a denoised action sequence:
\begin{equation}
    a^{ref}_{t:t+H^{ref}-1} = \pi^{\text{ref}}_{\theta}(\tilde{a}_{t:t+H^{ref}-1}, K^{\text{VLM}}_{1:N}, V^{\text{VLM}}_{1:N}).
\end{equation}
The generated sequence is then encoded via a MLP projector to obtain action embedding $Z^{\text{ex}}$, which serves as explicit action-space guidance $g_{\text{action}}^{\text{ex}}$ for action policy learning.

\subsection{Implicit Action Reasoner}
\label{sec:IAR}
Beyond the explicit action trajectories, the multimodal latent space of VLM also encodes implicit motion cues~\cite{qian2024affordancellm, driess2023palm}, \eg, visual affordances and action-related semantics. Effectively extracting these action-relevant representations potentially offers complementary guidance. To this end, we introduce an Implicit Action Reasoner (IAR), which directly operates on the VLM’s key–value cache.

Concretely, as presented in Fig.~\ref{fig:fig2} (b), for each VLM layer $i \in [1, N]$, we initialize a learnable matrix $Q_i \in \mathbb{R}^{M \times d}$, where $M$ is a hyperparameter and $d$ represents VLM’s hidden dimension.
Considering the information redundancy within VLM’s key–value cache and computational efficiency, we first downsample the corresponding key–value pairs into a lower-dimensional space, which is formulated as:
\begin{equation}
    Q_i' = Q_i W_Q^{(i)}, \quad 
    K_i' = K_i^{\text{VLM}} W_K^{(i)}, \quad 
    V_i' = V_i^{\text{VLM}} W_V^{(i)},
\end{equation}
where $W_Q^{(i)},W_K^{(i)},W_V^{(i)}\in \mathbb{R}^{d \times d'}$ are learnable linear projectors and $d'\ll d$.

Later, cross-attention is applied to extract action-relevant information from each $K_i'$ and $V_i'$.
The resulting features are subsequently integrated via average pooling and transformed through a MLP projector, as visualized in Fig.~\ref{fig:fig2} (b), producing compact representations that capture the implicit action semantics $z^{\text{im}}_i$ embedded in VLM's $i$-th layer:
\begin{equation}
    z^{\text{im}}_i = \text{MLP}(\text{Pool}(\text{CrossAttn}(Q_i', K_i', V_i'))).
\end{equation}
Then, through aggregating these representations across layers, we obtain implicit action-related feature $Z^{\text{im}}$, which serves as implicit action-space guidance $g_{\text{action}}^{\text{im}}$, complementing the explicit motion priors.

\begin{table*}[t]
    \centering
    \renewcommand{\arraystretch}{0.88}
    \resizebox{0.9\textwidth}{!}{
        \begin{tabular}{l|c|cc|cc|cc|cc|cc}
        \toprule
        \multirow{2}{*}{Methods} &
        \multirow{2}{*}{Guidance} &
        \multicolumn{2}{c|}{Spatial} &
        \multicolumn{2}{c|}{Object} &
        \multicolumn{2}{c|}{Goal} &
        \multicolumn{2}{c|}{Long} &
        \multicolumn{2}{c}{Avg.} \\[-0.8pt]
        \cmidrule(lr){3-12}
        & & SR~$\uparrow$ & Rank~$\downarrow$ &
            SR~$\uparrow$ & Rank~$\downarrow$ &
            SR~$\uparrow$ & Rank~$\downarrow$ &
            SR~$\uparrow$ & Rank~$\downarrow$ &
            SR~$\uparrow$ & Rank~$\downarrow$ \\[-1.0pt]
        \midrule

        Diffusion Policy~\cite{chi2025diffusion} & -- &
        78.3 & 26 & 92.5 & 18 & 68.3 & 27 & 50.5 & 27 & 72.4 & 27 \\
        Octo~\cite{team2024octo} & -- &
        78.9 & 25 & 85.7 & 26 & 84.6 & 20 & 51.1 & 26 & 75.1 & 25 \\
        \midrule
        
        CoT-VLA~\cite{zhao2025cot} & Visual &
        87.5 & 20 & 91.6 & 20 & 87.6 & 17 & 69.0 & 19 & 81.1 & 20 \\
        
        WorldVLA~\cite{cen2025worldvla}(256*256) & Visual &
        85.6 & 22 & 89.0 & 23 & 82.6 & 22 & 59.0 & 22 & 79.1 & 21 \\

        WorldVLA~\cite{cen2025worldvla}(512*512) & Visual &
        87.6 & 19 & 96.2 & 15 & 83.4 & 21 & 60.0 & 21 & 81.8 & 19 \\
        
        DreamVLA~\cite{zhang2025dreamvla} & Visual &
        97.5 & 9 & 94.0 & 16 & 89.5 & 15 & 89.5 & 12 & 92.6 & 14 \\
        
        UniVLA~\cite{wang2025unified} & Visual &
        95.4 & 14 & 98.8 & 4 & 93.6 & 12 & 94.0 & 6 & 95.5 & 10 \\

        F1~\cite{lv2025f1} & Visual &
        98.2 & 5 & 97.8 & 10 & 95.4 & 11 & 91.3 & 10 & 95.7 & 9 \\

        GE-Act~\cite{liao2025genie} & Visual &
        98.2 & 5 & 97.6 & 12 & 95.8 & 9 & 94.4 & 5 & 96.5 & 7 \\
        \midrule
        TraceVLA~\cite{zheng2024tracevla} & Linguistics &
        84.6 & 24 & 85.2 & 27 & 75.1 & 26 & 54.1 & 24 & 74.8 & 26 \\

        OpenVLA~\cite{kim2024openvla} & Linguistics &
        84.7 & 23 & 88.4 & 24 & 79.2 & 23 & 53.7 & 25 & 76.5 & 24 \\

        UniAct~\cite{zheng2025universal} & Linguistics &
        77.0 & 27 & 87.0 & 25 & 77.0 & 25 & 70.0 & 18 & 77.8 & 23 \\

        SpatialVLA~\cite{qu2025spatialvla} & Linguistics &
        88.2 & 18 & 89.9 & 22 & 78.6 & 24 & 55.5 & 23 & 78.1 & 22 \\

        ThinkAct~\cite{huang2025thinkact} & Linguistics &
        88.3 & 17 & 91.4 & 21 & 87.1 & 18 & 70.9 & 17 & 84.4 & 18 \\

        $\pi_{0}$-FAST~\cite{pertsch2025fast} & Linguistics &
        96.4 & 12 & 96.8 & 14 & 88.6 & 16 & 60.2 & 20 & 85.5 & 17 \\

        FPC-VLA~\cite{yang2025fpc} & Linguistics &
        87.0 & 21 & 92.0 & 19 & 86.2 & 19 & 82.2 & 15 & 86.9 & 16 \\

        SmolVLA~\cite{shukor2025smolvla} & Linguistics &
        93.0 & 16 & 94.0 & 16 & 91.0 & 14 & 77.0 & 16 & 88.8 & 15 \\
        
        GR00T-N1~\cite{bjorck2025gr00t} & Linguistics &
        94.4 & 15 & 97.6 & 12 & 93.0 & 13 & 90.6 & 11 & 93.9 & 13 \\

        $\pi_{0}$~\cite{black2410pi0} & Linguistics &
        96.8 & 11 & 98.8 & 4 & 95.8 & 9 & 85.2 & 14 & 94.1 & 12 \\

        GO-1~\cite{bu2025agibot} & Linguistics &
        96.2 & 13 & 97.8 & 10 & 96.0 & 8 & 89.2 & 13 & 94.8 & 11 \\

        DD-VLA~\cite{liang2025discrete} & Linguistics &
        97.2 & 10 & 98.6 & 6 & 97.4 & 5 & 92.0 & 9 & 96.3 & 8 \\

        MemoryVLA~\cite{shi2025memoryvla} & Linguistics &
        98.4 & 4 & 98.4 & 7 & 96.4 & 7 & 93.4 & 7 & 96.7 & 6 \\

        $\pi_{0.5}$~\cite{intelligence2025pi_} & Linguistics &
        98.8 & 2 & 98.2 & 9 & 98.0 & 3 & 92.4 & 8 & 96.9 & 5 \\

        OpenVLA-OFT~\cite{kim2025fine} & Linguistics &
        97.6 & 8 & 98.4 & 7 & 97.9 & 4 & 94.5 & 4 & 97.1 & 4 \\

        VLA-Adapter~\cite{wang2026vla} & Linguistics &
        97.8 & 7 & 99.2 & 2 & 97.2 & 6 & 95.0 & 3 & 97.3 & 3 \\

        \midrule
        \rowcolor{gray!20}
        \textbf{Ours$^{\diamond}$} & \textbf{Action} &
        \textbf{99.4} & \textbf{1} &
        \textbf{99.6} & \textbf{1} &
        98.8 & 2 &
        96.0 & 2 &
        \textbf{98.5} & \textbf{1} \\

        \rowcolor{gray!20}
        \textbf{Ours} & \textbf{Action} &
        98.6 & 3 &
        99.0 & 3 &
        \textbf{99.4} & \textbf{1} &
        \textbf{97.0} & \textbf{1} &
        \textbf{98.5} & \textbf{1} \\
        \bottomrule
    \end{tabular}
    }
    \vspace{-2mm}
    \caption{Comparison on the LIBERO benchmark. Our proposed approach is trained on the LIBERO dataset. ${\diamond}$ represents that the LLM backbone is frozen during training. All metrics are average success rates (\%). The best results are highlighted in \textbf{bold}.}
    \label{tab:libero}
    \vspace{-4.9mm}
\end{table*}

\subsection{Action-Guided Prediction}
\label{sec:AGP}
Building upon the explicit action embedding $Z^{\text{ex}}$ produced by EAR and implicit action-related feature $Z^{\text{im}}$ obtained in IAR, in this section, we introduce the Action-Guided Prediction (AGP) strategy to incorporate both action guidances into policy learning.

As illustrated in Fig.~\ref{fig:fig2} (c), given a noisy action segment $\tilde{a}_{t:t+H-1}$, we first encode it into noisy action embedding via a MLP projector. Particularly, unlike previous approaches that directly feed this embedding into action head $\pi_\theta^{\text{head}}$, we treat it as action query, denoted as $Q_{action}$, which interacts with both $Z^{\text{ex}}$ and $Z^{\text{im}}$ to retrieve complementary priors for conditional prediction.

Specifically, we perform dual cross-attention operations:
\vspace{-4mm}
\begin{align}
    S^{\text{ex}} &= \text{CrossAttn}(Q_{action}, Z^{\text{ex}}, Z^{\text{ex}}), \\
    S^{\text{im}} &= \text{CrossAttn}(Q_{action}, Z^{\text{im}}, Z^{\text{im}}),
\end{align}
where $S^{\text{ex}}$ and $S^{\text{im}}$ denote the attended representations guided by explicit and implicit priors, respectively.
Note that although both encode action-relevant information, they may highlight different facets of the underlying motion. For instance, explicit priors provide kinematic cues, whereas implicit priors capture latent action tendencies.
Hence, to effectively combine these complementary guidance, we concatenate the two attended features and process them through self-attention fusion block, which integrates the priors into a unified representation $\bar{h} $:
\begin{equation}
\bar{h} = \text{Self-Attn}([S^{\text{ex}};\, S^{\text{im}}]).
\end{equation}
Eventually, the aggregated representation $\bar{h}$ is fed into $\pi_\theta^{\text{head}}$, which predicts the denoised action sequence $a_{t:t+H-1}$.

\vspace{0.5mm}
\noindent\textbf{Training Objectives.}
The entire framework is optimized under a standard flow-matching mean-squared error (MSE) objective. The training losses consist of two parts, \ie, flow-matching MSE for both Explicit Action Reasoner $\pi_\theta^{\text{ref}}$ and action head $\pi_\theta^{\text{head}}$,  denoted as $\mathcal{L}_{\pi_\theta^{\text{ref}}}$ and $\mathcal{L}_{\pi_\theta^{\text{head}}}$, respectively. Hence, the overall objective is:
\begin{equation}
\mathcal{L}_{\text{total}} =
\lambda_{1} \mathcal{L}_{\pi_\theta^{\text{ref}}} +
\lambda_{2} \mathcal{L}_{\pi_\theta^{\text{head}}},
\end{equation}
where $\lambda_{1}$ and $\lambda_{2}$ are balance factors.

\vspace{0.5mm}
\noindent\textbf{Teacher Forcing Stabilization.}
During training, the outputs of $\pi_\theta^{\text{ref}}$ can be unstable. 
To stabilize optimization, we compute $Z^{\text{ex}}$ directly from ground-truth reference trajectories instead of from $\pi_\theta^{\text{ref}}$ predictions, preventing optimization interference to $\pi_\theta^{\text{head}}$. 
During inference, the model switches to a fully self-conditioned mode, 
where $\pi_\theta^{\text{ref}}$ autonomously generates the reference actions to guide $\pi_\theta^{\text{head}}$ in action prediction.
\begin{table*}[!t]
    \centering
    \renewcommand{\arraystretch}{0.88}
    \resizebox{0.9\textwidth}{!}{%
    \begin{tabular}{l|c|ccccccc|c}
        \toprule
        Methods & Guidance &
        Camera & Robot & Language & Light & Background & Noise & Layout & Avg. \\
        
        \midrule
        \midrule
        \multicolumn{10}{c}{\textit{Zero-Shot Transfer}} \\
        \midrule
        \midrule
        
        WorldVLA~\cite{cen2025worldvla} & Visual &
        0.1 & 27.9 & 41.6 & 43.7 & 17.1 & 10.9 & 38.0 & 25.0 \\
        OpenVLA~\cite{kim2024openvla} & Linguistics &
        0.8 & 3.5 & 23.0 & 8.1 & 34.8 & 15.2 & 28.5 & 15.6 \\
        NORA~\cite{hung2025nora} & Linguistics &
        2.2 & 37.0 & 65.1 & 45.7 & 58.6 & 12.8 & 62.1 & 39.0 \\
        UniVLA~\cite{bu2025univlalearningacttaskcentric} & Linguistics &
        1.8 & 46.2 & 69.6 & 69.0 & 81.0 & 21.2 & 31.9 & 42.9 \\
        $\pi_0$-Fast~\cite{pertsch2025fast} & Linguistics &
        65.1 & 21.6 & 61.0 & 73.2 & 73.2 & 74.4 & 68.8 & 61.6 \\
        RIPT-VLA~\cite{tan2025interactive} & Linguistics &
        55.2 & 31.2 & 77.6 & 88.4 & 91.6 & 73.5 & 74.2 & 68.4 \\
        OpenVLA-OFT~\cite{kim2025fine} & Linguistics &
        56.4 & 31.9 & 79.5 & 88.7 & 93.3 & 75.8 & 74.2 & 69.6 \\
        
        $\pi_0^*$~\cite{black2410pi0} & Linguistics & 61.0 & 40.8 & 63.5 & 89.3 & 84.1 & 80.1 & 76.4 & 69.4 \\
        $\pi_{0.5}^*$~\cite{intelligence2025pi_} & Linguistics & \textbf{75.8} & 79.4 & 83.3 & 95.5 & 95.0 & \textbf{89.6} & 87.0 & 85.7 \\

        \rowcolor{gray!20}
        \textbf{Ours$^\diamond$} & \textbf{Action} & 68.9 & 80.3 & 84.1 & 95.6 & 93.1 & 81.5 & \textbf{88.3} & 83.6 \\
        \rowcolor{gray!20}
        \textbf{Ours} & \textbf{Action} & 72.6 & \textbf{82.6} & \textbf{87.5} & \textbf{97.7} & \textbf{96.5} & 87.8 & 88.1 & \textbf{86.6} \\
        
        \midrule
        \midrule
        \multicolumn{10}{c}{\textit{Supervised Fine-Tuning}} \\
        \midrule
        \midrule
    
        $\pi_0^\diamond$~\cite{black2410pi0} & Linguistics & 79.6 & 21.1 & 72.5 & 84.7 & 86.2 & 68.3 & 69.4 & 67.4 \\
        $\pi_{0.5}^\diamond$~\cite{intelligence2025pi_} & Linguistics & 70.3 & 41.7 & \textbf{81.1} & \textbf{97.3} & 94.6 & 71.8 & 84.9 & 75.7 \\

        \rowcolor{gray!20}
        \textbf{Ours$^\diamond$} & \textbf{Action} & 91.2 & 62.5 & 80.3 & 95.1 & 91.5 & 88.3 & 84.9 & 84.1 \\
        \rowcolor{gray!20}
        \textbf{Ours} & \textbf{Action} & \textbf{96.6} & \textbf{70.4} & 79.7 & 95.1 & \textbf{97.1} & \textbf{95.9} & \textbf{85.0} & \textbf{88.0} \\
        \bottomrule
    \end{tabular}
    }
    \vspace{-2mm}
    \caption{Comparison on the LIBERO-Plus benchmark. Methods under \textit{Zero-Shot Transfer} are trained on LIBERO dataset and directly evaluated on LIBERO-Plus. \textit{Supervised Fine-Tuning} denotes models trained on the LIBERO-Plus training set. An asterisk (*) denotes results reproduced by utilizing officially released checkpoints, while $\diamond$ represents that the LLM backbone is frozen during training. The best results are highlighted in \textbf{bold}.}
    \label{tab:libero_plus}
    \vspace{-4.9mm}
\end{table*}
\section{Experiments}
\label{sec:exp}
In this section, we first outline the experimental setup in Sec.~\ref{sec:exp_setup}. Then, in Sec.~\ref{sec:exp_sim}, we evaluate our approach on three simulation benchmarks, followed by comprehensive ablation studies in Sec.~\ref{sec:exp_abla}. Moreover, we present real-world deployment results in Sec.~\ref{sec:exp_real} to evaluate real-world applicability.

\subsection{Experimental Setup}
\label{sec:exp_setup}
\noindent\textbf{Data Sources.}
For simulation experiments, we strictly follow the official training splits provided by the corresponding benchmark (LIBERO~\cite{liu2023libero}, LIBERO-Plus~\cite{fei2025libero}, and VLABench~\cite{zhang2025vlabench}), and train our models exclusively on their standard demonstration datasets without introducing any additional data.
For the real-world setting, all demonstrations used for model training are collected on our own robotic platform. More details about data sources are introduced in Appendix~\ref{appendix:dataset}.

\vspace{0.5mm}
\noindent\textbf{Implementation Details.}
We implement our approach upon $\pi_{0.5}$~\cite{intelligence2025pi_}. Specifically, we adopt SigLIP~\cite{zhai2023sigmoid} as the visual encoder, while the LLM backbone is instantiated as Gemma 2B architecture~\cite{beyer2024paligemma} with $N=18$ layers and hidden size $d=2048$. For frame processing, each input frame is resized to $224 \times 224$ prior to the visual encoder. Regarding the EAR, we employ a compact Transformer-based design composed of $N=18$ layers. Concerning the IAR, each learnable query matrix $Q_i$ is configured with a row dimension of $M=1$. The reduced dimension in the downsampling strategy is set to $d'=128$.

In terms of model training, unless explicitly specified, the horizon of predicted reference actions $H^{ref}$ and action policy output $H$ are fixed to $15$ and $10$, with action shift set to $2$ and $1$, respectively. To clarify, the action shift specifies the temporal interval relative to the expert demonstration. For instance, a shift of $1$ yields frame-aligned predictions, whereas a shift of $2$ skips one intermediate frame. We set the balance factors in training losses as $\lambda_1 =\lambda_2=0.5$.

\vspace{0.5mm}
\noindent\textbf{Training Configuration.}
We adopt a unified set of training hyperparameters across all experiments unless explicitly specified. Concretely, the learning rate follows a cosine-decay schedule with a warm-up phase of $10 \text{K}$ steps, a peak learning rate of $5\mathrm{e}{-5}$, and a decay toward $5\mathrm{e}{-5}$ over $10 \text{K}$ steps. Optimization is performed with AdamW with gradient-norm clipping set to $1.0$. An exponential moving average (EMA) of model parameters is maintained with a decay rate of $0.999$.
Regarding hardware settings, model training is performed on a single node equipped with $8$ NVIDIA H100 GPUs using bfloat16 precision. And the inference is conducted on a single NVIDIA RTX 4090.

\subsection{Simulation Experiments}
\label{sec:exp_sim}
In this section, we conduct the simulation evaluations across three benchmarks, \ie, LIBERO~\cite{liu2023libero}, LIBERO-Plus~\cite{fei2025libero}, and VLABench~\cite{zhang2025vlabench}, to comprehensively evaluate our approach's performance and generalization capabilities under diverse task structures.

\vspace{0.5mm}
\noindent\textbf{LIBERO Benchmark.}
We evaluate our approach on LIBERO benchmark, which targets four distinct robot capabilities: spatial awareness (Spatial), object manipulation (Object), goal completion (Goal), and long-horizon reasoning (Long).
Each task suite consists of $10$ tasks and provides $50$ human-teleoperated demonstrations per task for policy training.
The evaluation is conducted following the official evaluation protocol. For each task, the policy is evaluated over $50$ trials, amounting to $2,000$ total rollouts.

\begin{table*}[t]
    \centering
    \renewcommand{\arraystretch}{0.88}
    \small
    \resizebox{0.85\textwidth}{!}{%
    \begin{tabular}{l|c|cc|cc|cc|cc|cc|cc}
        \toprule
        \multirow{2}{*}{Methods} &
        \multirow{2}{*}{Guidance} &
        \multicolumn{2}{c|}{In-dist.} &
        \multicolumn{2}{c|}{Category} &
        \multicolumn{2}{c|}{Commonsense} &
        \multicolumn{2}{c|}{Instruction} &
        \multicolumn{2}{c|}{Texture} &
        \multicolumn{2}{c}{Avg.} \\[-0.6pt]

        \cmidrule(lr){3-14}

        & & IS~$\uparrow$ & PS~$\uparrow$ &
            IS~$\uparrow$ & PS~$\uparrow$ &
            IS~$\uparrow$ & PS~$\uparrow$ &
            IS~$\uparrow$ & PS~$\uparrow$ &
            IS~$\uparrow$ & PS~$\uparrow$ &
            IS~$\uparrow$ & PS~$\uparrow$ \\[-0.8pt]
        \midrule

        $\pi_{0}^\diamond$~\cite{black2410pi0} & Linguistics  &
        67.8 & 62.7 & 44.0 & 33.6 & 54.9 & \textbf{43.0} & \textbf{58.0} & 38.7 & 50.6 & 42.5 & 55.0 & 44.1  \\

        $\pi_{0.5}^\diamond$~\cite{intelligence2025pi_} & Linguistics  &
        75.0 & 60.8 & 49.6 & 35.3 & \textbf{57.5} & 41.6 & 57.1 & 30.3 & 62.0 & 47.4 & 60.2 & 43.1 \\
        \midrule

        \rowcolor{gray!20}
        \textbf{Ours$^\diamond$} & \textbf{Action} &
        \textbf{79.8} & \textbf{66.1} &
        \textbf{54.1} & \textbf{38.9} &
        52.3 & 37.8 &
        56.8 & \textbf{39.6} &
        \textbf{74.6} & \textbf{54.6} &
        \textbf{63.5} & \textbf{47.4} \\
        \bottomrule
    \end{tabular}
    }

    \vspace{-2.5mm}
    \caption{Comparison on the VLABench benchmark. IS and PS represent Intention score and Progress score, respectively. All models are trained for $60 \text{K}$ steps. ${\diamond}$ indicates that the LLM backbone is frozen during training. The best results are highlighted in \textbf{bold}.}
    \label{tab:vlabench}
    \vspace{-5mm}
\end{table*}

As reported in Table~\ref{tab:libero}, the quantitative evaluation results demonstrate that our proposed approach outperforms existing methods across all tracks. Compared to previous state-of-the-art method $\pi_{0.5}$, our approach achieves a $1.6\%$ absolute improvement in average.
Notably, we observe a pronounced improvement on LIBERO-Long suite, where tasks require long-horizon manipulation with strict error control. Particularly, unlike Language- or Visual-CoT, whose intermediate reasoning remains abstract or indirect with respect to action execution, our proposed ACoT naturally operates in precise representation. Through leveraging actions as intermediate reasoning, the model feeds the action head with structured action guidance, which significantly enhances the robustness in long-horizon manipulation tasks.

\begin{table}[!tp]
    \centering
    \renewcommand{\arraystretch}{0.92}
    \setlength{\tabcolsep}{8pt}
    \resizebox{\columnwidth}{!}{
    \begin{tabular}{c|cc|cccc|c}
        \toprule
        Name & EAR & IAR & Spatial & Object & Goal & Long & Avg. \\
        \midrule
        Baseline & & & 98.8 & 98.2 & 98.0 & 92.4 & 96.9 \\
        \midrule
        \#1 & \checkmark & & 99.0 & 99.4 & 98.0 & \textbf{96.6} & 98.3 \\
        \#2 & & \checkmark & 99.2 & 99.2 & 98.2 & 95.6 & 98.1 \\
        \#3 & \checkmark & \checkmark & \textbf{99.4} & \textbf{99.6} & \textbf{98.8} & 96.0 & \textbf{98.5} \\
        \bottomrule
    \end{tabular}
    }
    \vspace{-2mm}
    \caption{Module ablations. The performance is gradually improved with the continuous addition of proposed methods.}
    \label{tab:module_abla}
    \vspace{-6mm}
\end{table}
\vspace{0.5mm}
\noindent\textbf{LIBERO-Plus Benchmark.}
Built upon LIBERO, LIBERO-Plus is designed to systematically evaluate robotic policies under controlled distribution shifts. Concretely, LIBERO-Plus introduces $7$ perturbation dimensions, \ie, camera-viewpoints (Camera), robot-initial-states (Robot), language-variations (Language), lighting-conditions (Light), background-textures (Background), sensor-noise (Noise) and object-layout (Layout), exposing hidden failure modes under standard evaluations. Notably, LIBERO-Plus consists of $10,030$ evaluation episodes, providing statistically reliable evaluation.

We consider two evaluation protocols: (i) \textit{Zero-Shot Transfer}, where policies trained on the LIBERO dataset are directly evaluated on LIBERO-Plus to assess generalization. (ii) \textit{Supervised Fine-Tuning}, where models are directly optimized on LIBERO-Plus dataset. Technically, we follow evaluation configuration in LIBERO-Plus~\cite{fei2025libero}, \ie, each episode is executed once without repeated rollouts.

As shown in Table~\ref{tab:libero_plus}, our method significantly boosts the policy's performance, surpassing all previous methods in both settings.
Specifically, under the \textit{Zero-Shot} regime, our approach demonstrates pronounced robustness against distribution shifts such as robot initial-state perturbations ($+3.2\%$) and language variations ($+ 4.2\%$), where existing language-guided or vision-guided policies exhibit significant degradation.
Furthermore, our method maintains exceptional performance under the \textit{Supervised Fine-Tuning} setting, reaching an $88.0\%$ average success rate.
These results highlight the effectiveness of our action-space reasoning in improving generalization and robust policy learning.

\vspace{0.5mm}
\noindent\textbf{VLABench Benchmark.}
Built on ManiSkill3~\cite{tao2024maniskill3}, VLABench is designed to benchmark both VLAs and VLMs on diverse robotic tasks. The standard evaluation is organized into $5$ public tracks, \ie,
in-distribution, cross-category (category-level generalization), commonsense reasoning, semantic-instruction (language understanding), and unseen-texture (appearance robustness). Particularly, VLABench proposes Intention Score (IS) and Progress Score (PS) to evaluate robot policies.

In our context, we train $\pi_0$, $\pi_{0.5}$, along with our method in a unified training setup on VLABench's official training data. We present quantitative results in Table~\ref{tab:vlabench}. Overall, our method achieves the best performance across both IS  (${63.5\%}$) and PS (${47.4\%}$). Notably, under the challenging unseen-texture track, it delivers substantial gains, \ie, ${+12.6\%}$ in IS and ${+7.2\%}$ in PS, indicating strong robustness to distribution shifts. Together, these results further confirm the effectiveness of our proposed approach.

\begin{table}[!tp]
    \centering
    \renewcommand{\arraystretch}{1.0}
    \resizebox{\columnwidth}{!}{
    \begin{tabular}{c|c|c|c|cccc|c}
        \toprule
        Name & 
        \makecell{Action \\ shift} & 
        \makecell{Action \\ horizon} & 
        \makecell{Equi. \\ horizon} &
        Spatial & Object & Goal & Long & Avg. \\
        \midrule
        Baseline & 1 & 10 & 10 & 98.6 & 99.0 & 96.4 & 92.2 & 96.6 \\
        \midrule
         & 1 & 10 & 10 & 99.4 & 99.4 & \textbf{98.8} & 95.0 & 98.2 \\
         & 2 & 5 & 10 & \textbf{99.6} & \textbf{99.6} & 98.4 & 94.4 & 98.0 \\
         & 1 & 30 & 30 & 99.2 & 99.2 & 97.6 & 95.6 & 97.9 \\
        +EAR & 2 & 15 & 30 & 99.0 & 99.4 & 98.0 & \textbf{96.6} & \textbf{98.3} \\
         & 2 & 30 & 60 & 99.4 & 99.0 & 98.2 & 95.0 & 97.9 \\
         & 3 & 30 & 90 & 98.8 & 99.4 & 97.4 & 96.2 & 98.0 \\
        \bottomrule
    \end{tabular}
    }
    \vspace{-2mm}
    \caption{Reference action parameter ablation. We observe that different reference-action configurations within EAR generally lead to performance improvements.}
    \label{tab:ear_horizon}
    \vspace{-2mm}
\end{table}
\begin{table}[!tp]
    \centering
    \renewcommand{\arraystretch}{0.92}
    \setlength{\tabcolsep}{10pt}
    \resizebox{\columnwidth}{!}{
    \begin{tabular}{c|cccc|c}
        \toprule
        Methods & Spatial & Object & Goal & Long & Avg. \\
        \midrule
        Baseline & 98.8 & 98.2 & 98.0 & 92.4 & 96.9 \\
        \midrule
        Query & 98.8 & 99.0 & 97.2 & 92.8 & 97.0 \\
        Attention Pooling & \textbf{99.4} & 98.6 & \textbf{98.2} & 92.8 & 97.3 \\
        Downsample & 99.2 & \textbf{99.2} & \textbf{98.2} & \textbf{95.6} & \textbf{98.1} \\
        \bottomrule
    \end{tabular}
    }
    \vspace{-2mm}
    \caption{Comparison of KV-cache interaction strategies in IAR.}
    \label{tab:iar_sample}
    \vspace{-4mm}
\end{table}

\subsection{Ablation Study}
\label{sec:exp_abla}
We examine each component's contribution via systematic ablation experiments on the LIBERO benchmark, which are shown in Table~\ref{tab:module_abla}, Table~\ref{tab:ear_horizon}, and Table~\ref{tab:iar_sample}. Note that we adopt $\pi_{0.5}$ as the ``Baseline'' method. More ablations in different benchmarks are in Appendix~\ref{appendix:exps}.

\vspace{0.5mm}
\noindent\textbf{EAR.}
As shown in Table~\ref{tab:module_abla}, compared with the baseline, the experiment ``\#1'' introduces the Explicit Action Reasoner (EAR) module into policy learning, which lifts the average success rate from $96.9\%$ to $98.3\%$, demonstrating that the explicit action-space guidance benefits the robotic action sequence prediction.
A plausible explanation is that EAR introduces an intermediate reference action sequence, which injects strong inductive bias on the behavior and thereby reduces ambiguity in mapping from observations to actions.

\vspace{0.5mm}
\noindent\textbf{IAR.}
Analogously, with the  Implicit Action Reasoner (IAR) module added in ``\#2'', the average success rate increases from $96.9\%$ to $98.1\%$. This gain suggests that exploiting the implicit action distribution encoded in vision–language representations can also provide effective guidance for policy learning.
This performance gain can be partly attributed to the fact that IAR distills action-related clues implicitly encoded within the vision–language backbone, which potentially reflects the distribution of feasible actions. Such priors encourages the policy to remain closer to coherent, task-consistent behavioral patterns.

\vspace{0.5mm}
\noindent\textbf{EAR + IAR.}
In Table~\ref{tab:module_abla}, experiment ``\#3''  incorporates both EAR and IAR, achieving the highest average success rate of $98.5\%$. The consistent improvements demonstrate that explicit action guidance and implicit action cues extracted from VLM's key-value cache are complementary, jointly providing stronger guidance for accurate action prediction.

\vspace{0.5mm}
\noindent\textbf{Reference Action Configurations.}
To further examine the effect of explicit action references in EAR, we investigate different settings of action shift and action horizon, as summarized in Table~\ref{tab:ear_horizon}. We observe various parameter combinations consistently bring improvements over the baseline, indicating that providing action cues is broadly beneficial for policy learning. Besides, we find that shorter horizons combined with moderate shifts tend to produce relatively stronger gains. These observations offer further insight into how explicit action guidance influence policy learning.

\begin{figure}[t]
  \centering
  \vspace{-2mm}
  \includegraphics[width=\columnwidth]{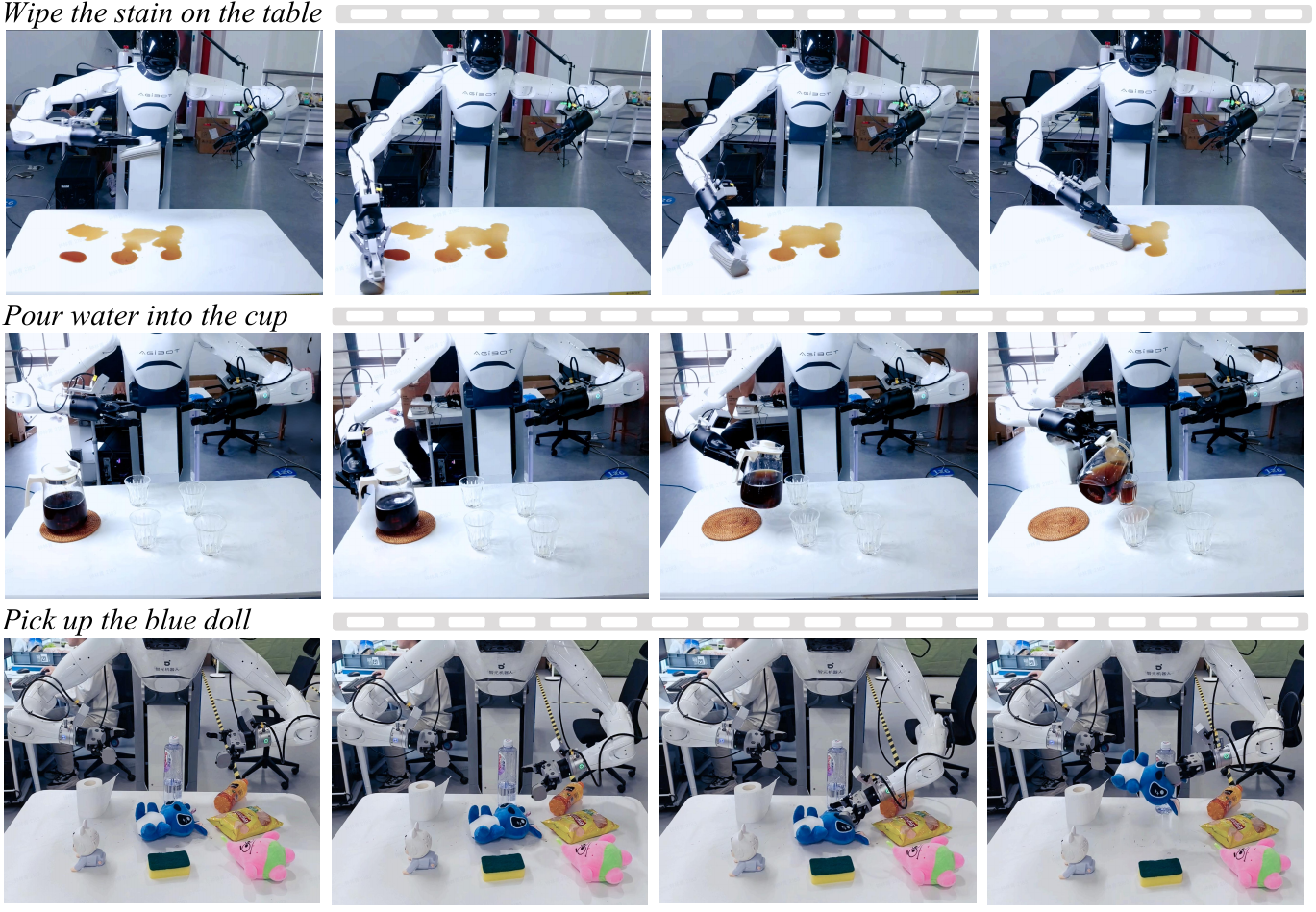}
  \vspace{-6mm}
  \caption{Visualization of three manipulation tasks in real world.}
  \vspace{-3mm}
  \label{fig:fig3}
\end{figure}

\vspace{0.5mm}
\noindent\textbf{KV-cache Interaction Strategies.}
We compare three strategies for extracting action-relevant cues from VLM's key-value cache within IAR module, as presented in Table~\ref{tab:iar_sample}.
Concretely, ``Query'' method uses learnable queries to attend to VLM's original key-value cache. 
``Attention Pooling'' method forms a pooled query by averaging key-value cache and then applies cross-attention operation. 
``Downsample'' method first downsamples VLM's key-value cache and then aggregates them using learnable matrix.

As shown in Table~\ref{tab:iar_sample}, all three variants outperform the baseline, indicating that extracting implicit action cues from VLM benefits policy learning. Notably, the ``Downsample'' strategy achieves the best performance, suggesting that VLM’s features may contain noisy information for action prediction. This also highlights the importance of designing appropriate interaction mechanisms to align vision-language and action.

\begin{figure}[t]
  \includegraphics[width=\linewidth]{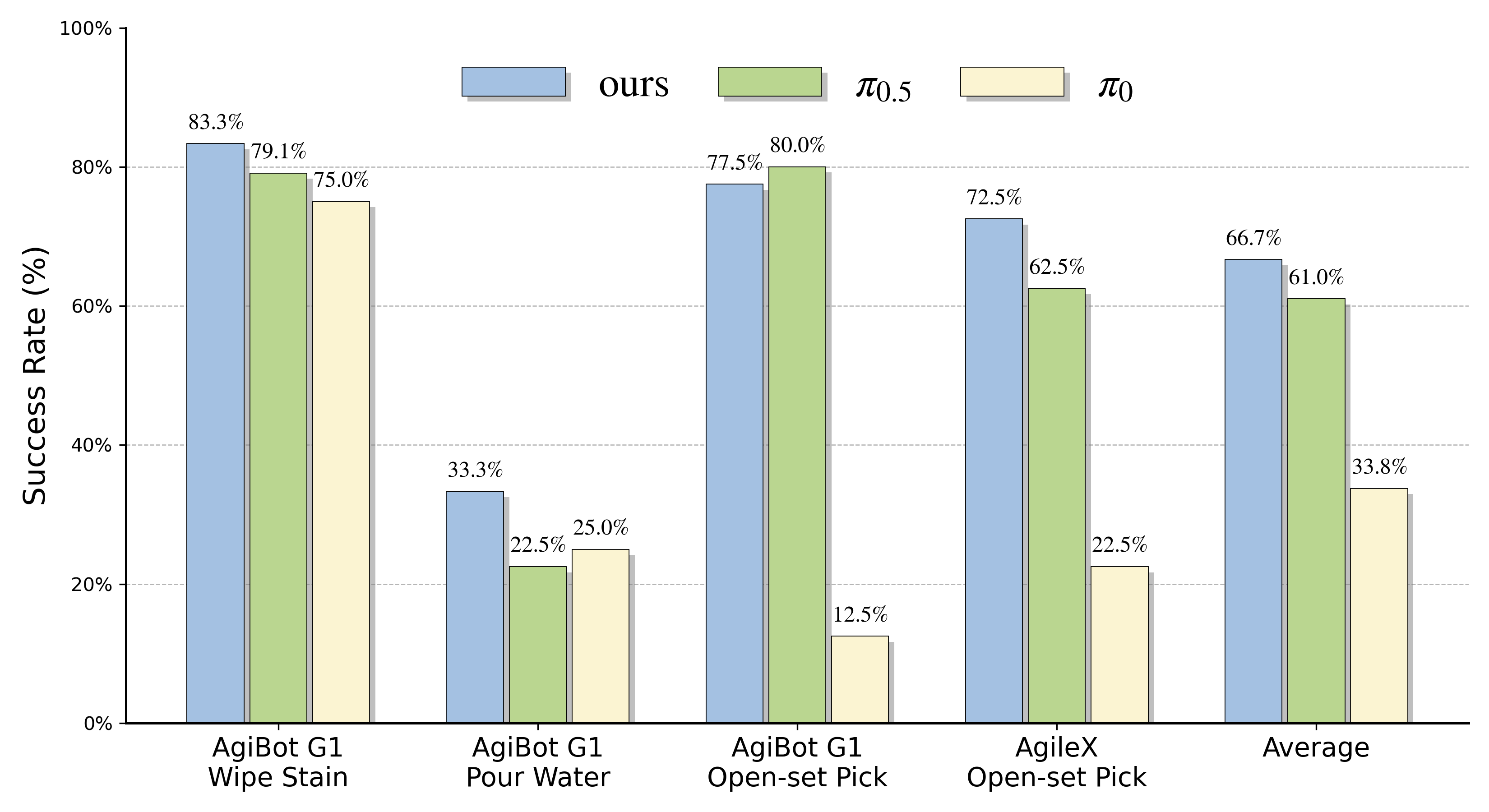}
  \vspace{-6mm}
  \caption{Evaluation results of real-world experiments.}
  \vspace{-3mm}
  \label{fig:fig4}
\end{figure}

\subsection{Real-World Deployment}
\label{sec:exp_real}
To further validate the effectiveness of our framework, we conduct extensive real-world experiments on the AgiBot G1 robot. We consider three manipulation tasks, \ie, ``Wipe Stain'', ``Pour Water'', and ``Open-set Pick'', which respectively assess contact-rich manipulation, fine-grained object handling, and instruction-following abilities.

Specifically, as visualized in Fig.~\ref{fig:fig3}, (i) the “Wipe Stain” task requires the robot to pick up a sponge from the table and wipe away the stain until the surface is clean. (ii) The “Pour Water” task requires the robot to grasp the kettle by its handle, locate the target cup, pour water into it without causing overflow, and finally return the kettle to the table in a stable manner. (iii) The “Open-set Pick” task instructs the robot to pick up the correct tabletop object according to given natural-language command. Additionally, to examine the cross-embodiment adaptability, we also perform the ``Open-set Pick'' task on the AgileX robotic platform. More details about training and evaluation are provided in the supplementary material.

As shown in Fig.~\ref{fig:fig4}, our approach achieves consistently higher average success rates than both $\pi_{0.5}$ and $\pi_0$, \ie, $66.7\%$ against $61.0\%$ and $33.8\%$. These results demonstrate that the proposed framework maintains effectiveness under real-world sensing conditions. Moreover, the aligned improvements observed on both AgiBot G1 and AgileX also indicate that our method exhibits adaptability across different robotic embodiments.
\section{Conclusion}
\label{sec:con}
In this work, we address the fundamental semantic-kinematic gap in modern robotic policies by proposing a new paradigm: Action Chain-of-Thought (ACoT). We argue that for physically grounded intelligence, deliberation should occur not in the abstract space of language or vision, but directly in the kinematically grounded space of actions. We materialize this concept in our ACoT-VLA framework, which leverages two synergistic modules---an Explicit Action Reasoner (EAR) and an Implicit Action Reasoner (IAR)---to generate and fuse both explicit trajectory plans and implicit behavioral priors. This action-centric guidance mechanism creates a direct, information-rich conduit between high-level intent and low-level motor control.
Extensive experiments across multiple simulation and real-world benchmarks demonstrate that our proposed approach yields state-of-the-art performance. Through shifting the locus of reasoning from perception to action, our work not only provides a more effective and grounded method for robotic policy learning, but also opens a new avenue for research into more structured, interpretable, and capable embodied agents. We believe that learning to ``think'' in the language of actions is a critical step towards developing the next generation of generalist robots.

{
    \small
    \bibliographystyle{ieeenat_fullname}
    \bibliography{main}

@String(AAAI = {AAAI})

@article{chi2025diffusion,
  title={Diffusion policy: Visuomotor policy learning via action diffusion},
  author={Chi, Cheng and Xu, Zhenjia and Feng, Siyuan and Cousineau, Eric and Du, Yilun and Burchfiel, Benjamin and Tedrake, Russ and Song, Shuran},
  journal={The International Journal of Robotics Research},
  volume={44},
  number={10-11},
  pages={1684--1704},
  year={2025},
  publisher={Sage Publications Sage UK: London, England}
}

@article{zhao2023learning,
  title={Learning fine-grained bimanual manipulation with low-cost hardware},
  author={Zhao, Tony Z and Kumar, Vikash and Levine, Sergey and Finn, Chelsea},
  journal={arXiv preprint arXiv:2304.13705},
  year={2023}
}

@article{team2024octo,
  title={Octo: An open-source generalist robot policy},
  author={Team, Octo Model and Ghosh, Dibya and Walke, Homer and Pertsch, Karl and Black, Kevin and Mees, Oier and Dasari, Sudeep and Hejna, Joey and Kreiman, Tobias and Xu, Charles and others},
  journal={arXiv preprint arXiv:2405.12213},
  year={2024}
}

@article{achiam2023gpt,
  title={Gpt-4 technical report},
  author={Achiam, Josh and Adler, Steven and Agarwal, Sandhini and Ahmad, Lama and Akkaya, Ilge and Aleman, Florencia Leoni and Almeida, Diogo and Altenschmidt, Janko and Altman, Sam and Anadkat, Shyamal and others},
  journal={arXiv preprint arXiv:2303.08774},
  year={2023}
}

@article{team2023gemini,
  title={Gemini: a family of highly capable multimodal models},
  author={Team, Gemini and Anil, Rohan and Borgeaud, Sebastian and Alayrac, Jean-Baptiste and Yu, Jiahui and Soricut, Radu and Schalkwyk, Johan and Dai, Andrew M and Hauth, Anja and Millican, Katie and others},
  journal={arXiv preprint arXiv:2312.11805},
  year={2023}
}

@article{bai2025qwen2,
  title={Qwen2. 5-vl technical report},
  author={Bai, Shuai and Chen, Keqin and Liu, Xuejing and Wang, Jialin and Ge, Wenbin and Song, Sibo and Dang, Kai and Wang, Peng and Wang, Shijie and Tang, Jun and others},
  journal={arXiv preprint arXiv:2502.13923},
  year={2025}
}

@article{kim2024openvla,
  title={Openvla: An open-source vision-language-action model},
  author={Kim, Moo Jin and Pertsch, Karl and Karamcheti, Siddharth and Xiao, Ted and Balakrishna, Ashwin and Nair, Suraj and Rafailov, Rafael and Foster, Ethan and Lam, Grace and Sanketi, Pannag and others},
  journal={arXiv preprint arXiv:2406.09246},
  year={2024}
}

@article{black2410pi0,
  title={$\pi_{0}$: A vision-language-action flow model for general robot control. CoRR, abs/2410.24164, 2024. doi: 10.48550},
  author={Black, Kevin and Brown, Noah and Driess, Danny and Esmail, Adnan and Equi, Michael and Finn, Chelsea and Fusai, Niccolo and Groom, Lachy and Hausman, Karol and Ichter, Brian and others},
  journal={arXiv preprint ARXIV.2410.24164}
}

@article{qu2025spatialvla,
  title={Spatialvla: Exploring spatial representations for visual-language-action model},
  author={Qu, Delin and Song, Haoming and Chen, Qizhi and Yao, Yuanqi and Ye, Xinyi and Ding, Yan and Wang, Zhigang and Gu, JiaYuan and Zhao, Bin and Wang, Dong and others},
  journal={arXiv preprint arXiv:2501.15830},
  year={2025}
}

@article{liu2024rdt,
  title={Rdt-1b: a diffusion foundation model for bimanual manipulation},
  author={Liu, Songming and Wu, Lingxuan and Li, Bangguo and Tan, Hengkai and Chen, Huayu and Wang, Zhengyi and Xu, Ke and Su, Hang and Zhu, Jun},
  journal={arXiv preprint arXiv:2410.07864},
  year={2024}
}

@article{ha2018world,
  title={World models},
  author={Ha, David and Schmidhuber, J{\"u}rgen},
  journal={arXiv preprint arXiv:1803.10122},
  volume={2},
  number={3},
  year={2018}
}

@inproceedings{yan2025renderworld,
  title={Renderworld: World model with self-supervised 3d label},
  author={Yan, Ziyang and Dong, Wenzhen and Shao, Yihua and Lu, Yuhang and Liu, Haiyang and Liu, Jingwen and Wang, Haozhe and Wang, Zhe and Wang, Yan and Remondino, Fabio and others},
  booktitle={2025 IEEE International Conference on Robotics and Automation (ICRA)},
  pages={6063--6070},
  year={2025},
  organization={IEEE}
}

@inproceedings{zheng2024occworld,
  title={Occworld: Learning a 3d occupancy world model for autonomous driving},
  author={Zheng, Wenzhao and Chen, Weiliang and Huang, Yuanhui and Zhang, Borui and Duan, Yueqi and Lu, Jiwen},
  booktitle={European conference on computer vision},
  pages={55--72},
  year={2024},
  organization={Springer}
}

@article{zhen20243d,
  title={3d-vla: A 3d vision-language-action generative world model},
  author={Zhen, Haoyu and Qiu, Xiaowen and Chen, Peihao and Yang, Jincheng and Yan, Xin and Du, Yilun and Hong, Yining and Gan, Chuang},
  journal={arXiv preprint arXiv:2403.09631},
  year={2024}
}

@article{zhang2025dreamvla,
  title={Dreamvla: a vision-language-action model dreamed with comprehensive world knowledge},
  author={Zhang, Wenyao and Liu, Hongsi and Qi, Zekun and Wang, Yunnan and Yu, Xinqiang and Zhang, Jiazhao and Dong, Runpei and He, Jiawei and Lu, Fan and Wang, He and others},
  journal={arXiv preprint arXiv:2507.04447},
  year={2025}
}

@inproceedings{zitkovich2023rt,
  title={Rt-2: Vision-language-action models transfer web knowledge to robotic control},
  author={Zitkovich, Brianna and Yu, Tianhe and Xu, Sichun and Xu, Peng and Xiao, Ted and Xia, Fei and Wu, Jialin and Wohlhart, Paul and Welker, Stefan and Wahid, Ayzaan and others},
  booktitle={Conference on Robot Learning},
  pages={2165--2183},
  year={2023},
  organization={PMLR}
}

@inproceedings{peebles2023scalable,
  title={Scalable diffusion models with transformers},
  author={Peebles, William and Xie, Saining},
  booktitle={Proceedings of the IEEE/CVF international conference on computer vision},
  pages={4195--4205},
  year={2023}
}

@article{zhou2024transfusion,
  title={Transfusion: Predict the next token and diffuse images with one multi-modal model},
  author={Zhou, Chunting and Yu, Lili and Babu, Arun and Tirumala, Kushal and Yasunaga, Michihiro and Shamis, Leonid and Kahn, Jacob and Ma, Xuezhe and Zettlemoyer, Luke and Levy, Omer},
  journal={arXiv preprint arXiv:2408.11039},
  year={2024}
}

@article{lipman2022flow,
  title={Flow matching for generative modeling},
  author={Lipman, Yaron and Chen, Ricky TQ and Ben-Hamu, Heli and Nickel, Maximilian and Le, Matt},
  journal={arXiv preprint arXiv:2210.02747},
  year={2022}
}

@article{bu2025agibot,
  title={Agibot world colosseo: A large-scale manipulation platform for scalable and intelligent embodied systems},
  author={Bu, Qingwen and Cai, Jisong and Chen, Li and Cui, Xiuqi and Ding, Yan and Feng, Siyuan and Gao, Shenyuan and He, Xindong and Hu, Xuan and Huang, Xu and others},
  journal={arXiv preprint arXiv:2503.06669},
  year={2025}
}

@article{deng2025graspvla,
  title={Graspvla: a grasping foundation model pre-trained on billion-scale synthetic action data},
  author={Deng, Shengliang and Yan, Mi and Wei, Songlin and Ma, Haixin and Yang, Yuxin and Chen, Jiayi and Zhang, Zhiqi and Yang, Taoyu and Zhang, Xuheng and Zhang, Wenhao and others},
  journal={arXiv preprint arXiv:2505.03233},
  year={2025}
}

@article{zhang2025spatial,
  title={From Spatial to Actions: Grounding Vision-Language-Action Model in Spatial Foundation Priors},
  author={Zhang, Zhengshen and Li, Hao and Dai, Yalun and Zhu, Zhengbang and Zhou, Lei and Liu, Chenchen and Wang, Dong and Tay, Francis EH and Chen, Sijin and Liu, Ziwei and others},
  journal={arXiv preprint arXiv:2510.17439},
  year={2025}
}

@article{yuan2025depthvla,
  title={DepthVLA: Enhancing Vision-Language-Action Models with Depth-Aware Spatial Reasoning},
  author={Yuan, Tianyuan and Liu, Yicheng and Lu, Chenhao and Chen, Zhuoguang and Jiang, Tao and Zhao, Hang},
  journal={arXiv preprint arXiv:2510.13375},
  year={2025}
}

@article{zhang20254d,
  title={4D-VLA: Spatiotemporal Vision-Language-Action Pretraining with Cross-Scene Calibration},
  author={Zhang, Jiahui and Chen, Yurui and Xu, Yueming and Huang, Ze and Zhou, Yanpeng and Yuan, Yu-Jie and Cai, Xinyue and Huang, Guowei and Quan, Xingyue and Xu, Hang and others},
  journal={arXiv preprint arXiv:2506.22242},
  year={2025}
}

@article{lin2025evo,
  title={Evo-0: Vision-language-action model with implicit spatial understanding},
  author={Lin, Tao and Li, Gen and Zhong, Yilei and Zou, Yanwen and Du, Yuxin and Liu, Jiting and Gu, Encheng and Zhao, Bo},
  journal={arXiv preprint arXiv:2507.00416},
  year={2025}
}

@article{duan2024manipulate,
  title={Manipulate-anything: Automating real-world robots using vision-language models},
  author={Duan, Jiafei and Yuan, Wentao and Pumacay, Wilbert and Wang, Yi Ru and Ehsani, Kiana and Fox, Dieter and Krishna, Ranjay},
  journal={arXiv preprint arXiv:2406.18915},
  year={2024}
}

@article{huang2024a3vlm,
  title={A3vlm: Actionable articulation-aware vision language model},
  author={Huang, Siyuan and Chang, Haonan and Liu, Yuhan and Zhu, Yimeng and Dong, Hao and Gao, Peng and Boularias, Abdeslam and Li, Hongsheng},
  journal={arXiv preprint arXiv:2406.07549},
  year={2024}
}

@article{huang2024rekep,
  title={Rekep: Spatio-temporal reasoning of relational keypoint constraints for robotic manipulation},
  author={Huang, Wenlong and Wang, Chen and Li, Yunzhu and Zhang, Ruohan and Fei-Fei, Li},
  journal={arXiv preprint arXiv:2409.01652},
  year={2024}
}

@inproceedings{li2024manipllm,
  title={Manipllm: Embodied multimodal large language model for object-centric robotic manipulation},
  author={Li, Xiaoqi and Zhang, Mingxu and Geng, Yiran and Geng, Haoran and Long, Yuxing and Shen, Yan and Zhang, Renrui and Liu, Jiaming and Dong, Hao},
  booktitle={Proceedings of the IEEE/CVF Conference on Computer Vision and Pattern Recognition},
  pages={18061--18070},
  year={2024}
}

@article{intelligence2025pi_,
  title={$\pi_{0.5}$: a Vision-Language-Action Model with Open-World Generalization},
  author={Intelligence, Physical and Black, Kevin and Brown, Noah and Darpinian, James and Dhabalia, Karan and Driess, Danny and Esmail, Adnan and Equi, Michael and Finn, Chelsea and Fusai, Niccolo and others},
  journal={arXiv preprint arXiv:2504.16054},
  year={2025}
}

@article{team2025gemini,
  title={Gemini robotics: Bringing ai into the physical world},
  author={Team, Gemini Robotics and Abeyruwan, Saminda and Ainslie, Joshua and Alayrac, Jean-Baptiste and Arenas, Montserrat Gonzalez and Armstrong, Travis and Balakrishna, Ashwin and Baruch, Robert and Bauza, Maria and Blokzijl, Michiel and others},
  journal={arXiv preprint arXiv:2503.20020},
  year={2025}
}

@article{huang2025motvla,
  title={MoTVLA: A Vision-Language-Action Model with Unified Fast-Slow Reasoning},
  author={Huang, Wenhui and Chen, Changhe and Qi, Han and Lv, Chen and Du, Yilun and Yang, Heng},
  journal={arXiv preprint arXiv:2510.18337},
  year={2025}
}

@inproceedings{o2024open,
  title={Open x-embodiment: Robotic learning datasets and rt-x models: Open x-embodiment collaboration 0},
  author={O’Neill, Abby and Rehman, Abdul and Maddukuri, Abhiram and Gupta, Abhishek and Padalkar, Abhishek and Lee, Abraham and Pooley, Acorn and Gupta, Agrim and Mandlekar, Ajay and Jain, Ajinkya and others},
  booktitle={2024 IEEE International Conference on Robotics and Automation (ICRA)},
  pages={6892--6903},
  year={2024},
  organization={IEEE}
}

@article{chen2025robotwin,
  title={Robotwin 2.0: A scalable data generator and benchmark with strong domain randomization for robust bimanual robotic manipulation},
  author={Chen, Tianxing and Chen, Zanxin and Chen, Baijun and Cai, Zijian and Liu, Yibin and Li, Zixuan and Liang, Qiwei and Lin, Xianliang and Ge, Yiheng and Gu, Zhenyu and others},
  journal={arXiv preprint arXiv:2506.18088},
  year={2025}
}

@inproceedings{zhang2025vlabench,
  title={Vlabench: A large-scale benchmark for language-conditioned robotics manipulation with long-horizon reasoning tasks},
  author={Zhang, Shiduo and Xu, Zhe and Liu, Peiju and Yu, Xiaopeng and Li, Yuan and Gao, Qinghui and Fei, Zhaoye and Yin, Zhangyue and Wu, Zuxuan and Jiang, Yu-Gang and others},
  booktitle={Proceedings of the IEEE/CVF International Conference on Computer Vision},
  pages={11142--11152},
  year={2025}
}

@article{jiang2025galaxea,
  title={Galaxea open-world dataset and g0 dual-system vla model},
  author={Jiang, Tao and Yuan, Tianyuan and Liu, Yicheng and Lu, Chenhao and Cui, Jianning and Liu, Xiao and Cheng, Shuiqi and Gao, Jiyang and Xu, Huazhe and Zhao, Hang},
  journal={arXiv preprint arXiv:2509.00576},
  year={2025}
}

@article{zhang2025up,
  title={Up-vla: A unified understanding and prediction model for embodied agent},
  author={Zhang, Jianke and Guo, Yanjiang and Hu, Yucheng and Chen, Xiaoyu and Zhu, Xiang and Chen, Jianyu},
  journal={arXiv preprint arXiv:2501.18867},
  year={2025}
}

@article{lv2025f1,
  title={F1: A vision-language-action model bridging understanding and generation to actions},
  author={Lv, Qi and Kong, Weijie and Li, Hao and Zeng, Jia and Qiu, Zherui and Qu, Delin and Song, Haoming and Chen, Qizhi and Deng, Xiang and Pang, Jiangmiao},
  journal={arXiv preprint arXiv:2509.06951},
  year={2025}
}

@article{li2025vla,
  title={Vla-rft: Vision-language-action reinforcement fine-tuning with verified rewards in world simulators},
  author={Li, Hengtao and Ding, Pengxiang and Suo, Runze and Wang, Yihao and Ge, Zirui and Zang, Dongyuan and Yu, Kexian and Sun, Mingyang and Zhang, Hongyin and Wang, Donglin and others},
  journal={arXiv preprint arXiv:2510.00406},
  year={2025}
}

@inproceedings{zhao2025cot,
  title={Cot-vla: Visual chain-of-thought reasoning for vision-language-action models},
  author={Zhao, Qingqing and Lu, Yao and Kim, Moo Jin and Fu, Zipeng and Zhang, Zhuoyang and Wu, Yecheng and Li, Zhaoshuo and Ma, Qianli and Han, Song and Finn, Chelsea and others},
  booktitle={Proceedings of the Computer Vision and Pattern Recognition Conference},
  pages={1702--1713},
  year={2025}
}

@article{cen2025worldvla,
  title={WorldVLA: Towards Autoregressive Action World Model},
  author={Cen, Jun and Yu, Chaohui and Yuan, Hangjie and Jiang, Yuming and Huang, Siteng and Guo, Jiayan and Li, Xin and Song, Yibing and Luo, Hao and Wang, Fan and others},
  journal={arXiv preprint arXiv:2506.21539},
  year={2025}
}

@article{liao2025genie,
  title={Genie envisioner: A unified world foundation platform for robotic manipulation},
  author={Liao, Yue and Zhou, Pengfei and Huang, Siyuan and Yang, Donglin and Chen, Shengcong and Jiang, Yuxin and Hu, Yue and Cai, Jingbin and Liu, Si and Luo, Jianlan and others},
  journal={arXiv preprint arXiv:2508.05635},
  year={2025}
}

@article{chen2022analog,
  title={Analog bits: Generating discrete data using diffusion models with self-conditioning},
  author={Chen, Ting and Zhang, Ruixiang and Hinton, Geoffrey},
  journal={arXiv preprint arXiv:2208.04202},
  year={2022}
}

@article{liu2022self,
  title={Self-conditioned generative adversarial networks for image editing},
  author={Liu, Yunzhe and Gal, Rinon and Bermano, Amit H and Chen, Baoquan and Cohen-Or, Daniel},
  journal={arXiv preprint arXiv:2202.04040},
  year={2022}
}

@article{wang2025unified,
  title={Unified Vision-Language-Action Model},
  author={Wang, Yuqi and Li, Xinghang and Wang, Wenxuan and Zhang, Junbo and Li, Yingyan and Chen, Yuntao and Wang, Xinlong and Zhang, Zhaoxiang},
  journal={arXiv preprint arXiv:2506.19850},
  year={2025}
}

@article{zheng2024tracevla,
  title={Tracevla: Visual trace prompting enhances spatial-temporal awareness for generalist robotic policies},
  author={Zheng, Ruijie and Liang, Yongyuan and Huang, Shuaiyi and Gao, Jianfeng and Daum{\'e} III, Hal and Kolobov, Andrey and Huang, Furong and Yang, Jianwei},
  journal={arXiv preprint arXiv:2412.10345},
  year={2024}
}

@article{huang2025thinkact,
  title={Thinkact: Vision-language-action reasoning via reinforced visual latent planning},
  author={Huang, Chi-Pin and Wu, Yueh-Hua and Chen, Min-Hung and Wang, Yu-Chiang Frank and Yang, Fu-En},
  journal={arXiv preprint arXiv:2507.16815},
  year={2025}
}

@article{yang2025fpc,
  title={FPC-VLA: A Vision-Language-Action Framework with a Supervisor for Failure Prediction and Correction},
  author={Yang, Yifan and Duan, Zhixiang and Xie, Tianshi and Cao, Fuyu and Shen, Pinxi and Song, Peili and Jin, Piaopiao and Sun, Guokang and Xu, Shaoqing and You, Yangwei and others},
  journal={arXiv preprint arXiv:2509.04018},
  year={2025}
}

@article{shukor2025smolvla,
  title={Smolvla: A vision-language-action model for affordable and efficient robotics},
  author={Shukor, Mustafa and Aubakirova, Dana and Capuano, Francesco and Kooijmans, Pepijn and Palma, Steven and Zouitine, Adil and Aractingi, Michel and Pascal, Caroline and Russi, Martino and Marafioti, Andres and others},
  journal={arXiv preprint arXiv:2506.01844},
  year={2025}
}

@article{bjorck2025gr00t,
  title={Gr00t n1: An open foundation model for generalist humanoid robots},
  author={Bjorck, Johan and Casta{\~n}eda, Fernando and Cherniadev, Nikita and Da, Xingye and Ding, Runyu and Fan, Linxi and Fang, Yu and Fox, Dieter and Hu, Fengyuan and Huang, Spencer and others},
  journal={arXiv preprint arXiv:2503.14734},
  year={2025}
}

@article{shi2025memoryvla,
  title={Memoryvla: Perceptual-cognitive memory in vision-language-action models for robotic manipulation},
  author={Shi, Hao and Xie, Bin and Liu, Yingfei and Sun, Lin and Liu, Fengrong and Wang, Tiancai and Zhou, Erjin and Fan, Haoqiang and Zhang, Xiangyu and Huang, Gao},
  journal={arXiv preprint arXiv:2508.19236},
  year={2025}
}

@inproceedings{zheng2025universal,
  title={Universal actions for enhanced embodied foundation models},
  author={Zheng, Jinliang and Li, Jianxiong and Liu, Dongxiu and Zheng, Yinan and Wang, Zhihao and Ou, Zhonghong and Liu, Yu and Liu, Jingjing and Zhang, Ya-Qin and Zhan, Xianyuan},
  booktitle={Proceedings of the Computer Vision and Pattern Recognition Conference},
  pages={22508--22519},
  year={2025}
}

@article{kim2025fine,
  title={Fine-tuning vision-language-action models: Optimizing speed and success},
  author={Kim, Moo Jin and Finn, Chelsea and Liang, Percy},
  journal={arXiv preprint arXiv:2502.19645},
  year={2025}
}

@article{liang2025discrete,
  title={Discrete diffusion vla: Bringing discrete diffusion to action decoding in vision-language-action policies},
  author={Liang, Zhixuan and Li, Yizhuo and Yang, Tianshuo and Wu, Chengyue and Mao, Sitong and Pei, Liuao and Yang, Xiaokang and Pang, Jiangmiao and Mu, Yao and Luo, Ping},
  journal={arXiv preprint arXiv:2508.20072},
  year={2025}
}

@article{pertsch2025fast,
  title={Fast: Efficient action tokenization for vision-language-action models},
  author={Pertsch, Karl and Stachowicz, Kyle and Ichter, Brian and Driess, Danny and Nair, Suraj and Vuong, Quan and Mees, Oier and Finn, Chelsea and Levine, Sergey},
  journal={arXiv preprint arXiv:2501.09747},
  year={2025}
}

@article{hung2025nora,
  title={Nora: A small open-sourced generalist vision language action model for embodied tasks},
  author={Hung, Chia-Yu and Sun, Qi and Hong, Pengfei and Zadeh, Amir and Li, Chuan and Tan, U and Majumder, Navonil and Poria, Soujanya and others},
  journal={arXiv preprint arXiv:2504.19854},
  year={2025}
}

@misc{bu2025univlalearningacttaskcentric,
      title={UniVLA: Learning to Act Anywhere with Task-centric Latent Actions}, 
      author={Qingwen Bu and Yanting Yang and Jisong Cai and Shenyuan Gao and Guanghui Ren and Maoqing Yao and Ping Luo and Hongyang Li},
      year={2025},
      eprint={2505.06111},
      archivePrefix={arXiv},
      primaryClass={cs.RO},
      url={https://arxiv.org/abs/2505.06111}, 
}

@inproceedings{qian2024affordancellm,
  title={Affordancellm: Grounding affordance from vision language models},
  author={Qian, Shengyi and Chen, Weifeng and Bai, Min and Zhou, Xiong and Tu, Zhuowen and Li, Li Erran},
  booktitle={Proceedings of the IEEE/CVF Conference on Computer Vision and Pattern Recognition},
  pages={7587--7597},
  year={2024}
}

@article{driess2023palm,
  title={Palm-e: An embodied multimodal language model},
  author={Driess, Danny and Xia, Fei and Sajjadi, Mehdi SM and Lynch, Corey and Chowdhery, Aakanksha and Wahid, Ayzaan and Tompson, Jonathan and Vuong, Quan and Yu, Tianhe and Huang, Wenlong and others},
  year={2023}
}

@article{liu2023libero,
  title={Libero: Benchmarking knowledge transfer for lifelong robot learning},
  author={Liu, Bo and Zhu, Yifeng and Gao, Chongkai and Feng, Yihao and Liu, Qiang and Zhu, Yuke and Stone, Peter},
  journal={Advances in Neural Information Processing Systems},
  volume={36},
  pages={44776--44791},
  year={2023}
}

@article{fei2025libero,
  title={LIBERO-Plus: In-depth Robustness Analysis of Vision-Language-Action Models},
  author={Fei, Senyu and Wang, Siyin and Shi, Junhao and Dai, Zihao and Cai, Jikun and Qian, Pengfang and Ji, Li and He, Xinzhe and Zhang, Shiduo and Fei, Zhaoye and others},
  journal={arXiv preprint arXiv:2510.13626},
  year={2025}
}

@inproceedings{zhai2023sigmoid,
  title={Sigmoid loss for language image pre-training},
  author={Zhai, Xiaohua and Mustafa, Basil and Kolesnikov, Alexander and Beyer, Lucas},
  booktitle={Proceedings of the IEEE/CVF international conference on computer vision},
  pages={11975--11986},
  year={2023}
}

@article{beyer2024paligemma,
  title={Paligemma: A versatile 3b vlm for transfer},
  author={Beyer, Lucas and Steiner, Andreas and Pinto, Andr{\'e} Susano and Kolesnikov, Alexander and Wang, Xiao and Salz, Daniel and Neumann, Maxim and Alabdulmohsin, Ibrahim and Tschannen, Michael and Bugliarello, Emanuele and others},
  journal={arXiv preprint arXiv:2407.07726},
  year={2024}
}

@article{zawalski2024robotic,
  title={Robotic control via embodied chain-of-thought reasoning},
  author={Zawalski, Micha{\l} and Chen, William and Pertsch, Karl and Mees, Oier and Finn, Chelsea and Levine, Sergey},
  journal={arXiv preprint arXiv:2407.08693},
  year={2024}
}

@article{tao2024maniskill3,
  title={Maniskill3: Gpu parallelized robotics simulation and rendering for generalizable embodied ai},
  author={Tao, Stone and Xiang, Fanbo and Shukla, Arth and Qin, Yuzhe and Hinrichsen, Xander and Yuan, Xiaodi and Bao, Chen and Lin, Xinsong and Liu, Yulin and Chan, Tse-kai and others},
  journal={arXiv preprint arXiv:2410.00425},
  year={2024}
}

@article{tan2025interactive,
  title={Interactive Post-Training for Vision-Language-Action Models},
  author={Tan, Shuhan and Dou, Kairan and Zhao, Yue and Kr{\"a}henb{\"u}hl, Philipp},
  journal={arXiv preprint arXiv:2505.17016},
  year={2025}
}

@article{liu2025unified,
  title={Unified Embodied VLM Reasoning with Robotic Action via Autoregressive Discretized Pre-training},
  author={Liu, Yi and Wang, Sukai and Wei, Dafeng and Cai, Xiaowei and Zhong, Linqing and Yang, Jiange and Ren, Guanghui and Zhang, Jinyu and Yao, Maoqing and Li, Chuankang and others},
  journal={arXiv preprint arXiv:2512.24125},
  year={2025}
}

@inproceedings{wang2026vla,
  title={Vla-adapter: An effective paradigm for tiny-scale vision-language-action model},
  author={Wang, Yihao and Ding, Pengxiang and Li, Lingxiao and Cui, Can and Ge, Zirui and Tong, Xinyang and Song, Wenxuan and Zhao, Han and Zhao, Wei and Hou, Pengxu and others},
  booktitle={Proceedings of the AAAI conference on artificial intelligence},
  volume={40},
  number={22},
  pages={18638--18646},
  year={2026}
}

@article{yin2026genie,
  title={Genie Sim 3.0: A High-Fidelity Comprehensive Simulation Platform for Humanoid Robot},
  author={Yin, Chenghao and Huang, Da and Yang, Di and Wang, Jichao and Zhao, Nanshu and Xu, Chen and Sun, Wenjun and Hou, Linjie and Li, Zhijun and Wu, Junhui and others},
  journal={arXiv preprint arXiv:2601.02078},
  year={2026}
}
}

\clearpage
\appendix
\renewcommand{\thesection}{\Alph{section}}
\section{Dataset Description}
\label{appendix:dataset}
In this section, we present a comprehensive characterization of the benchmark datasets and the custom-collected data used for model training in our experiments. We systematically report key statistics, including the total number of episodes, frame counts, and other relevant properties, which is summarized in Table~\ref{tab:dataset_statistics} below:
\begin{table}[h]
    \centering
    \renewcommand{\arraystretch}{1.0}
    \setlength{\tabcolsep}{6pt}
    \resizebox{\columnwidth}{!}{
    \begin{tabular}{l|c|ccccc}
        \toprule
        Type & Dataset & Embodiment & DoF & Episodes & Frames & FPS \\
        \midrule
        \multirow{3}{*}{Simulation} & LIBERO & Franka & 7 & 1,693 & 273,465 & 10 \\
        & LIBERO-Plus & Franka & 7 & 14,347 & 2,238,036 & 20 \\
        & VLABench & Franka & 7 & 4,713 & 528,398 & 10 \\
        \midrule
        \multirow{4}{*}{Real-World} & Wipe Stain & AgiBot G1 & 22 & 177 & 356,316 & 30 \\
        & Pour Water & AgiBot G1 & 22 & 1,821 & 5,062,506 & 30 \\
        & Open-set Pick & AgiBot G1 & 22 & 1,936 & 219,824 & 30 \\
        & Open-set Pick & AgileX & 14 & 962 & 251,283 & 30 \\
        \bottomrule
    \end{tabular}
    }
    \vspace{-2mm}
    \caption{Dataset statistics.}
    \label{tab:dataset_statistics}
    \vspace{-4mm}
\end{table}

\noindent\textbf{Simulation Benchmarks.}
We utilize three publicly released simulation datasets, \ie, LIBERO~\cite{liu2023libero}, LIBERO-Plus~\cite{fei2025libero}, and VLABench~\cite{zhang2025vlabench}.
Specifically, the LIBERO dataset contains $1,693$ episodes and $273,465$ frames, recorded at a fixed $10$ Hz. Its demonstrations exhibit relatively uniform trajectory lengths and smooth motion patterns, making it widely adopted benchmark in community.

However, due to the increasing performance saturation observed on LIBERO, LIBERO-Plus is recently introduced to provide a more challenging and diversified evaluation setting. LIBERO-Plus provides $14,347$ episodes and $2,238,036$ frames, captured at $20$ Hz.
In contrast to the homogeneous trajectories in LIBERO, LIBERO-Plus explicitly emphasizes a perturbation-oriented design. The demonstrations display substantially larger variations in motion magnitude and camera–robot viewpoint configuration.
These characteristics make it a more suitable benchmark for evaluating policy generalization under structured distribution shifts.

Besides these two datasets, we further benchmark our method on VLABench, whose training set includes $4,713$ episodes and $528,398$ frames, recorded at $10$ Hz, which requires a higher level of visual and physical understanding from the policy.

\vspace{0.5mm}
\noindent\textbf{Real-World Experiment.}
For real-world deployment, we collect demonstrations across $3$ tasks, \ie, Wipe Stain, Pour Water, and Open-set Pick, as shown in Table~\ref{tab:dataset_statistics}.

The ``Wipe Stain'' dataset contains $177$ episodes with $356,316$ frames, characterized by dense tool–surface contact and fine-grained force control. The ``Pour Water'' dataset includes $1,821$ episodes and $5,062,506$ frames. Its large scale stems from the task’s long-horizon and multi-stage nature. Regarding the ``Open-set Pick'' task, the AgiBot G1 subset provides $1,936$ episodes with $219,824$ frames, while the AgileX subset offers $962$ episodes with $251,283$ frames, both featuring diverse tabletop layouts and natural-language instructions.

\section{Training \& Evaluation Details}
\label{appendix:train_eval_details}
\begin{table}[t]
    \centering
    \renewcommand{\arraystretch}{1.0}
    \setlength{\tabcolsep}{6pt}
    \resizebox{\columnwidth}{!}{
    \begin{tabular}{l|ccccc}
        \toprule
        Task & Action Space & Action Horizon & State & Batch Size & Training Step \\
        \midrule
        LIBERO & Delta EEF & 10 & $\times$ & 128 & 40K \\
        LIBERO-Plus & Delta EEF & 10 & $\times$ & 128 & 100K \\
        VLABench & Abs EEF & 10 & \checkmark & 128 & 60K \\
        \midrule
        Wipe Stain & Abs Joint & 30 & \checkmark & 128 & 50K \\
        Pour Water & Abs Joint & 30 & \checkmark & 128 & 240K \\
        Open-set Pick & Abs Joint & 30 & \checkmark & 128 & 50K \\
        Open-set Pick$^{\dagger}$ & Abs Joint & 30 & \checkmark & 128 & 50K \\
        \bottomrule
    \end{tabular}
    }
    \vspace{-2mm}
    \caption{Training details. Note that the ``Open-set Pick$^{\dagger}$'' task is performed on AgileX platform.}
    \label{tab:training_details}
    \vspace{-4mm}
\end{table}
\noindent\textbf{Training Details.}
We describe the task-specific training configurations, \eg, action space and state usage, for better understanding.

As presented in Table~\ref{tab:training_details}, for the LIBERO and LIBERO-Plus suites, the policy is trained using delta end-effector control (Delta EEF) with an action horizon of $10$ steps. In particular, no privileged state information is provided during training. We utilize a global batch size of $128$ and train the policies for $40\text{K}$ and $100\text{K}$ steps, respectively. Similarly, we train our models in VLABench for $60\text{K}$ steps, while adopting state input and absolute end-effector (Abs EEF) actions to align the benchmark’s control convention.

In terms of the real-world tasks, we utilize Abs Joint control with a longer action horizon of $30$. 
Unlike the simulator benchmarks, these tasks additionally provide structured robot state observations to improve robustness under real-world sensing and actuation noise.
Our models are trained for $50\text{K}$, $240\text{K}$, and $50\text{K}$ steps, in ``Wipe Stain'', ``Pour Water'', and ``Open-set Pick'' tasks, respectively, with same batch size of $128$.
\begin{table*}[!tp]
    \centering
    \renewcommand{\arraystretch}{1.0}
    \resizebox{0.92\textwidth}{!}{
    \begin{tabular}{c|cc|ccccccc|c}
        \toprule
        Name & EAR & IAR & Camera & Robot & Language & Light & Background & Noise & Layout & Avg. \\
        \midrule
        Baseline & & & 70.3 & 41.7 & 81.1 & 97.3 & \textbf{94.6} & 71.8 & 84.9 & 75.7 \\
        \midrule
        \#1 & \checkmark & & 88.7 & \textbf{63.5} & 80.4 & 94.0 & 90.2 & \textbf{89.5} & 84.2 & 83.7 \\
        \#2 & & \checkmark & 80.7 & 48.7 & \textbf{82.6} & \textbf{97.7} & 90.9 & 84.3 & \textbf{86.0} & 80.4 \\
        \#3 & \checkmark & \checkmark & \textbf{91.2} & 62.5 & 80.3 & 95.1 & 91.5 & 88.3 & 84.9 & \textbf{84.1} \\
        \bottomrule
    \end{tabular}
    }
    \vspace{-2mm}
    \caption{Module ablations on LIBERO-Plus benchmark. The performance is gradually improved with the addition of proposed methods. Note that models are directly optimized on LIBERO-Plus dataset, with the LLM backbone frozen during training.}
    \label{tab:module_abla_libero_plus}
\end{table*}
\begin{table*}[!tp]
    \centering
    \renewcommand{\arraystretch}{1.25}
    \setlength{\tabcolsep}{3pt}
    \resizebox{\textwidth}{!}{
    \begin{tabular}{c|cc|cc|ccccc|cccccccc}
        \toprule
        \multirow{2}{*}{Name}
        & \multicolumn{2}{c|}{Action Head} & \multicolumn{2}{c|}{EAR} 
        & \multicolumn{5}{c|}{LIBERO}
        & \multicolumn{8}{c}{LIBERO-Plus} \\

        & Param. & Denoise 
        & Param. & Denoise
        & Spatial & Object & Goal & Long & Avg.
        & Camera & Robot & Language & Light & Background & Noise & Layout & Avg. \\

        \midrule
        Baseline & 300M & 10 & - & - 
        & 98.6 & 99.0 & 96.4 & 92.2 & 96.6
        & 70.3 & 41.7 & 81.1 & \underline{97.3} & \textbf{94.6} & 71.8 & \underline{84.9} & 75.7 \\
        \midrule

        \#1 & 600M & 10 & - & - 
        & 97.6 & 98.4 & 97.8 & \underline{96.4} & 97.6
        & 68.7 & 44.8 & \underline{83.1} & 96.4 & 92.7 & 66.6 & 84.1 & 74.9 \\

        \#2 & 600M & 20 & - & -
        & 97.8 & 98.8 & \underline{98.0} & 95.2 & 97.5
        & 70.0 & 44.8 & 82.7 & \textbf{97.6} & 93.1 & 66.7 & 83.2 & 75.1 \\

        \#3 & 300M & 5 & 300M & 5
        & 98.6 & \textbf{99.6} & 97.8 & 95.4 & \underline{97.9}
        & \underline{88.2} & \underline{62.4} & 81.5 & 95.0 & 91.5 & 88.6 & \textbf{85.3} & \textbf{83.9} \\

        \#4 & 300M & 10 & 300M & 10
        & \underline{99.0} & \underline{99.4} & \underline{98.0} & \textbf{96.6} & \textbf{98.3}
        & \textbf{88.7} & \textbf{63.5} & 80.4 & 94.0 & 90.2 & \textbf{89.5} & 84.2 & \underline{83.7} \\
        \midrule
        \midrule

        \#4 & 300M & 10 & 300M & 10
        & \underline{99.0} & \underline{99.4} & \underline{98.0} & \textbf{96.6} & \textbf{98.3}
        & \textbf{88.7} & \textbf{63.5} & 80.4 & 94.0 & 90.2 & \textbf{89.5} & 84.2 & \underline{83.7} \\

        \#5 & 300M & 10 & 150M & 10
        & \textbf{99.2} & 99.2 & 97.8 & 94.2 & 97.6
        & 86.4 & 54.3 & 81.7 & 92.2 & 91.4 & \underline{89.1} & 82.1 & 81.7 \\

        \#6 & 300M & 10 & 250M & 10
        & \underline{99.0} & 98.2 & \textbf{98.6} & 94.2 & 97.5
        & 87.2 & 59.7 & 81.1 & 95.0 & \underline{93.7} & 87.4 & 83.5 & 83.1 \\

        \#7 & 300M & 10 & 500M & 10
        & 98.4 & \underline{99.4} & 96.6 & 94.2 & 97.0
        & 80.8 & 57.6 & \textbf{84.1} & 95.6 & 92.1 & 79.8 & 83.7 & 80.9 \\

        \bottomrule
    \end{tabular}
    }
    \vspace{-2mm}
    \caption{Effects of parameters and denoise steps on policy performance. Note that the IAR module is not added in this experiment. The evaluation protocol in LIBERO-Plus is \textit{Supervised Fine-Tuning}, \ie, models are directly optimized on LIBERO-Plus dataset. The LLM backbone is frozen during training. The best results are highlighted in \textbf{bold}, and the second-best results are \underline{underlined}.}
    \label{tab:ear_param_denoise}
    \vspace{-4mm}
\end{table*}

\vspace{0.5mm}
\noindent\textbf{Evaluation Details.}
Next, we illustrate the evaluation protocols and success criteria for all real-world tasks. Each task is assessed using fixed and repeatable initializations to ensure reproducibility and reduce environmental variance.

Concretely, in terms of the ``Wipe Stain'' task, we predefine three initial sponge poses. For each pose, the robot is required to clean stains placed at four distinct table locations. Every configuration is executed twice, resulting in $24$ trials in total. A trial is considered successful if the robot grasps the sponge and removes the stain from the specified location.

As for the ``Pour Water'', we standardize six predefined relative configurations between the bottle and the glass. Then, each configuration is executed two times. A trial is counted as successful if the robot lifts the bottle, pours water into the cup, and places the bottle back onto the coaster. Note that minor spillage of water when pouring is allowed.

Eventually, regarding the ``Open-set Pick'' task, we initialize ten object arrangements on the table, containing both in-distribution and out-of-distribution instances. In each arrangement, the robot is instructed to grasp a specified target object using either its left or right arm, as indicated by the instruction. Each arm–object pair is evaluated twice, resulting in $40$ trials overall. A trial is deemed successful if the robot grasps the instructed object with the correct arm.

Across all tasks, evaluations are carried out by trained operators with substantial prior testing experience, and success rates are computed as the proportion of successful trials relative to the total number of executed attempts.

\section{More Experimental Results}
\label{appendix:exps}
In this section, we provide additional quantitative experiments to substantiate the effectiveness of our proposed approach and to empirically uncover several insightful phenomena. Specifically, the experimental analyses comprise four parts: (1) ablation study conducted on the LIBERO-Plus benchmark in Table~\ref{tab:module_abla_libero_plus}, (2) an investigation of how the parameter sizes of the Action Head and Explicit Action Reasoner (EAR), as well as the number of denoising steps, influence policy performance in Table~\ref{tab:ear_param_denoise}, (3) evaluation of proposed approach's sim2real capability based on Genie-Sim 3.0~\cite{yin2026genie} in Table~\ref{tab:sim2real}, and (4) a comparative study examining the relationship among inference latency, model size, and performance in Table~\ref{tab:efficiency}. Note that we adopt $\pi_{0.5}$~\cite{intelligence2025pi_} as the baseline method, denoted as ``Baseline''.

\vspace{0.5mm}
\noindent\textbf{Module Ablation.}
As shown in Table~\ref{tab:module_abla_libero_plus}, incorporating the proposed reasoning modules consistently improves policy performance on the LIBERO-Plus benchmark.
Adding the EAR module, \ie, experiment ``\#1'', yields a clear gain over the baseline, increasing the average success rate from $75.7\%$ to $83.7\%$. This improvement can be attributed to EAR's ability to generate an explicit reference action trajectory, which significantly reduces the ambiguity in mapping complex visual or linguistic observations to low-level actions, such as camera shifts and background changes.
Meanwhile, incorporating only the IAR (``\#2'') also improves the performance from $75.7\%$ to $80.4\%$, indicating that decoding the latent action-related semantics within the vision–language backbone provides useful behavioral priors.
Finally, combining EAR and IAR (``\#3'') achieves the highest success rate of $84.1\%$, demonstrating their complementary effects, \ie, EAR provides explicit motion guidance, while IAR supplies dense representation-level priors.

\vspace{0.5mm}
\noindent\textbf{Effect of Model Scaling \& Denoising Budget.}
Then, we analyze the superiority of our method by comparing settings with matched total model parameters and denoising steps. As shown in Table~\ref{tab:ear_param_denoise}, firstly, we enlarge the model size of the action head and increase the number of denoising steps in experiments ``\#1'' and ``\#2'', to construct fair baselines for subsequent comparison.
We observe a preliminary observation, \ie, increasing the model size or denoising steps does not reliably enhance performance. Specifically, compared with the baseline, while ``\#1'' improves performance on the LIBERO benchmark, it simultaneously drops on LIBERO-Plus. Next, comparing ``\#1'' and ``\#2'' reveals that further increasing denoising steps yields only negligible fluctuations.

Subsequently, we incorporate the EAR module under fully matched overall parameterization and denoising budgets. Concretely, in both comparison pairs, ``\#1'' with ``\#3'' and ``\#2'' with ``\#4'', we consistently observe notable performance improvements on both benchmarks, once the EAR module is introduced. This indicates that the performance gains originate from our proposed action chain-of-thought. The proposed mechanism supplies explicit reference actions that effectively mitigate the intrinsic instability of action prediction, especially under challenging external perturbations, as shown in the LIBERO-Plus, enabling a more reliable and grounded generalist robotic policy.

\vspace{0.5mm}
\noindent\textbf{Effect of EAR Scale.}
Moreover, we investigate how various scale of the EAR module influences action prediction fidelity. To isolate the effect of EAR, we keep the action head parameters and the denoising schedule strictly fixed, while scaling the EAR module to $150\text{M}$, $250\text{M}$, $300\text{M}$, and $500\text{M}$ parameters via adjusting hidden size.
As presented in Table~\ref{tab:ear_param_denoise}, through the comparison across experiments ``\#4'', ``\#5'', ``\#6'', and ``\#7'', we find that although all EAR-equipped variants outperform non-EAR baselines on both benchmarks, the performance trend is non-monotonic. Applying moderate EAR scales, \eg, $300\text{M}$, yields the greatest improvement.
Particularly, as evidenced in ``\#7'' in Table~\ref{tab:ear_param_denoise}, when the parameter of EAR module even exceeds that of the action head, we observe a marked drop in performance. We attribute this degradation to the tendency of an over-parameterized EAR to overfit spurious correlations during training. Therefore, it generates reference action trajectories that are systematically biased, which ultimately misdirect the action head toward suboptimal predictions.

\begin{table}[!t]
    \centering
    \renewcommand{\arraystretch}{0.9}
    \resizebox{0.8\columnwidth}{!}{%
    \begin{tabular}{l|cc|cc}
        \toprule
        \multirow{2}{*}{Tasks} & \multicolumn{2}{c|}{Simulation} & \multicolumn{2}{c}{Real-World} \\
        \cmidrule(lr){2-3} \cmidrule(lr){4-5}
        & $\pi_{0.5}$ & Ours & $\pi_{0.5}$ & Ours \\
        \midrule
        Select Color         & 86.0 & \textbf{98.8} & 85.0 & \textbf{94.0} \\
        Recognize Size       & 93.0 & \textbf{96.0} & \textbf{94.0} & \textbf{94.0} \\
        Grasp Targets        & \textbf{71.7} & 68.0 & 70.8 & \textbf{75.0} \\
        Organize Objects     & 52.0 & \textbf{74.0} & 60.0 & \textbf{68.4} \\
        \midrule
        Avg.                 & 75.7 & \textbf{84.2} & 77.5 & \textbf{82.9} \\
        \bottomrule
    \end{tabular}
    }
    \vspace{-2mm}
    \caption{Experimental results on Genie Sim 3.0 Simulation and Real-world Transfer. The best results are highlighted in \textbf{bold}.}
    \label{tab:sim2real}
    \vspace{-6mm}
\end{table}

\vspace{0.5mm}
\noindent\textbf{Sim To Real Evaluation.}
To further assess the sim2real transferability of proposed method, we conduct evaluations on the Genie-Sim 3.0 benchmark~\cite{yin2026genie}. It comprises $4$ challenging tasks: picking objects based on specified colors, sizes, or categories, as well as a complex tabletop organization task. Technically, the model is trained on the officially released simulation-based datasets. Then, it is deployed and evaluated in both simulation and real-world environments.

As illustrated in Table~\ref{tab:sim2real}, our approach consistently outperforms the baseline across both domains. Specifically, it achieves $84.2\%$ in simulation and $82.9\%$ in real-world settings, representing absolute improvements of $8.5\%$ and $5.4\%$, respectively. Notably, our method exhibits minimal performance degradation during the sim-to-real transition, which we attribute to the fundamental nature of ACoT paradigm. Specifically, while visual domain gap persists between synthetic and real-world observations, the underlying kinematically grounded action guidance remains consistent. By shifting the locus of reasoning from the perceptual space to the action space, our model extracts task-relevant motion priors that are invariant to low-level visual perturbations, effectively enhancing policy's sim2real capability.

\begin{table}[!tp]
    \centering
    \renewcommand{\arraystretch}{1.0}
    \setlength{\tabcolsep}{6pt}
    \resizebox{\columnwidth}{!}{
    \begin{tabular}{c|cc|cc|cc}
        \toprule
        \multirow{2}{*}{Name} & \multirow{2}{*}{EAR} & \multirow{2}{*}{IAR} & \multirow{2}{*}{Param.} & \multirow{2}{*}{Latency} & LIBERO & LIBERO-Plus \\
        & & & & & Avg. SR & Avg. SR \\
        \midrule
        Baseline & & & 3.35B & 91ms & 96.9 & 75.7 \\
        \midrule
        \#1 & \checkmark & & 3.80B & 110ms & 98.3 & 83.7 \\
        \#2 & & \checkmark & 3.36B & 93ms & 98.1 & 80.4 \\
        \#3 & \checkmark & \checkmark & 3.81B & 112ms & \textbf{98.5} & \textbf{84.1} \\
        \bottomrule
    \end{tabular}
    }
    \vspace{-2mm}
    \caption{Ablation experiment on model efficiency and performance. Note that the evaluation protocol in LIBERO-Plus is \textit{Supervised Fine-Tuning}.}
    \label{tab:efficiency}
    \vspace{-4mm}
\end{table}

\vspace{0.5mm}
\noindent\textbf{Latency Analysis.}
In Table~\ref{tab:efficiency}, we further examine the inference efficiency of our approach in terms of both parameter count and end-to-end latency. As additional reasoning modules are introduced, we observe a slight increase in latency. Incorporating the EAR module raises latency from $91\text{ms}$ to $110\text{ms}$, while adding the IAR module introduces only an additional $2\text{ms}$. However, this marginal overhead is outweighed by the substantial improvement, which reflects a favorable trade-off.

\section{Limitations \& Future Works}
In this section, we discuss the limitations existing in our work and promising directions for future research.

Although our proposed action chain-of-thought (ACoT) substantially boosts policy performance, our framework still exhibits several constraints. The reasoning modules introduce additional computational cost, which, while relatively modest compared to the performance gains, may pose challenges for deployment on resource-constrained robotic platforms.
Besides, another limitation stems from the fact that the prevailing action representation in the community is implemented as action chunks, \ie, sequences of low-level control commands such as joint angles or end-effector poses. While such representations faithfully encode the executed motions, they lack explicit geometric structure that would facilitate higher-level spatial reasoning, such as object-centric coordination and contact geometry. Hence, the potential of ACoT reasoning may not be fully unleashed. Enriching action representations with spatially grounded information to enable ACoT to operate in geometrically interpretable 3D space, constitutes an interesting and promising avenue for future exploration.

\section{LLM Usage Statement}
In this paper, we employ Large Language Models (LLMs) solely for minor linguistic refinement during the manuscript preparation stage, such as correcting grammatical errors. None of the technical content, implementation details, or experimental results were generated by LLMs.
\end{document}